\documentclass[oneside,a4paper,onecolumn,11pt]{article}

\usepackage{fullpage}
\usepackage[left=2cm,top=2cm,bottom=2cm,right=2cm,includehead,nomarginpar,headheight=16pt]{geometry}
\usepackage[ruled, linesnumbered, vlined, commentsnumbered]{algorithm2e}
\usepackage{graphicx,subfig} % figure related
\usepackage{amsfonts,amssymb,amsmath,amsthm,amsopn,mathtools}	% math related
\usepackage{booktabs,diagbox,colortbl,multirow,tabularx,threeparttable,hhline}
\usepackage[listings,skins,breakable]{tcolorbox}
\usepackage{algorithmic}
\usepackage{fancyhdr,fancyheadings,nopageno,lastpage} % setting header and footer
\usepackage{enumerate}
\usepackage[shortlabels]{enumitem}
\usepackage{csquotes}
\usepackage{authblk}
\usepackage{footnote}
\usepackage[colorlinks=true,citecolor=reference,linkcolor=brightmaroon,backref=page]{hyperref}
\usepackage{prettyref}
\usepackage{cite}
\usepackage{setspace}
\usepackage{color}
\usepackage{titlesec}
\usepackage{soul}
\usepackage{url}
\usepackage{bm}
\usepackage{xcolor}  % Required for custom colors
\usepackage{csquotes}
\usepackage{academicons}
\usepackage{fontawesome5}
\usepackage{wrapfig}
\usepackage{pifont,ifsym,marvosym,manfnt} % math fonts
\usepackage[framemethod=TikZ]{mdframed}

\setlist[itemize]{itemsep=0pt}
\setlist[enumerate]{itemsep=0pt}

\renewcommand{\headrulewidth}{0pt} % Remove line at top
\fancypagestyle{plain}{%
    \renewcommand{\headrulewidth}{0pt}%
    \fancyfoot[C]{}
    \fancyfoot[R]{\faLeanpub\ \, \thepage\ / \pageref{LastPage}}
    \fancyfoot[L]{\faBraille\ \textsc{COLALab Report} \faSlackHash\ \footnotesize\textifsym{2023}}
}

\hypersetup{
    colorlinks=true,
    linkcolor=ultramarine,
    filecolor=magenta,
    urlcolor=ultramarine,
    pdftitle={Overleaf Example},
    pdfpagemode=FullScreen,
}

\definecolor{reference}{RGB}{4, 20, 110}
\definecolor{amaranth}{rgb}{0.9, 0.17, 0.31}
\definecolor{brightmaroon}{rgb}{0.76, 0.13, 0.28}
\definecolor{specialgreen}{rgb}{0.5176, 0.8275, 0.2392}
\definecolor{specialblue}{rgb}{0.3412, 0.6745, 0.9412}
\definecolor{deepblue}{rgb}{0.2118, 0.2784, 0.6039}
\definecolor{rowcolor}{RGB}{96,54,148}
\definecolor{lightpurple}{RGB}{242, 239, 246}
\definecolor{ultramarine}{RGB}{0,32,96}

\newrefformat{fig}{Fig.~\ref{#1}}
\newrefformat{tab}{Table~\ref{#1}}
\newrefformat{sec}{Section~\ref{#1}}
\newrefformat{app}{Appendix~\ref{#1}}
\newrefformat{alg}{Algorithm~\ref{#1}}
\newrefformat{remark}{Remark~\ref{#1}}

\newtcolorbox{quotebox}{
    colback=lightpurple,
    colframe=black!75,
    boxrule=0pt,
    top=5pt,
    bottom=5pt,
    left=8pt,
    right=8pt,
    arc=8pt,
    boxsep=0pt,
    toptitle=2pt,
    bottomtitle=2pt,
    fonttitle=\bfseries,
}

\definecolor{NavyBlue}{RGB}{8,29,92}

\definecolor{set_blue}{RGB}{116,159,201}
\definecolor{set_light_blue}{RGB}{215,229,242}

\definecolor{set_light_orange}{RGB}{247,222,190}
\definecolor{set_mid_orange}{RGB}{244,197,154}
\definecolor{set_orange}{RGB}{243,174,123}

\definecolor{set_green}{RGB}{128,184,134}
\definecolor{set_green_light}{RGB}{194,226,184}
\definecolor{set_green_mid}{RGB}{219,238,211}

\definecolor{set_grey}{RGB}{138,138,138}
\definecolor{set_mid_grey}{RGB}{206,206,206}
\definecolor{set_light_grey}{RGB}{227,227,227}

\definecolor{set_purple}{RGB}{151,146,191}
\definecolor{set_mid_purple}{RGB}{183,180,211}
\definecolor{set_light_purple}{RGB}{228,228,239}

\newsavebox\newcaptionbox\newdimen\newcaptionboxwid
\titlespacing*{\paragraph}{0pt}{0.75ex plus 0.75ex minus 0.2ex}{1.25ex plus 0.2ex}

\long\def\@makecaption#1#2{
 \vskip 10pt
        \baselineskip 11pt
        \setbox\@tempboxa\hbox{#1. #2}
        \ifdim \wd\@tempboxa >\hsize
        \sbox{\newcaptionbox}{\small\sl #1.~}
        \newcaptionboxwid=\wd\newcaptionbox
        \usebox\newcaptionbox {\footnotesize #2}
        \else
          \centerline{{\small\sl #1.} {\small #2}}
        \fi}

\SetCommentSty{mycommfont}

\newcommand{\pref}{\prettyref}

\newrefformat{fig}{Fig.~\ref{#1}}
\newrefformat{tab}{Table~\ref{#1}}
\newrefformat{sec}{Section~\ref{#1}}
\newrefformat{alg}{Algorithm~\ref{#1}}
\newrefformat{property}{Property~\ref{#1}}
\newrefformat{theorem}{Theorem~\ref{#1}}
\newrefformat{definition}{Definition~\ref{#1}}
\newrefformat{corollary}{Corollary~\ref{#1}}
\newrefformat{lemma}{Lemma~\ref{#1}}
\newrefformat{conj}{Conjecture~\ref{#1}}
\newrefformat{def}{Definition~\ref{#1}}
\newrefformat{eq}{equation~(\ref{#1})}
\newrefformat{app}{Appendix~\ref{#1}}

\usepackage{lscape}

\newenvironment{code-example}
{
\vspace{0.15cm}
\noindent\begin{minipage}{\linewidth}
\begin{center}
\arrayrulecolor{black}
\color{black}
\begin{tabular}{|p{0.95\linewidth}|}
\hline%
\rowcolor{pink!20}%
}
{
\\\hline
\end{tabular}
\end{center}
\end{minipage}
\vspace{-0.2cm}
}

\begin{document}

%% title
\title{\vspace{-1ex}\LARGE\textbf{On the Hyperparameter Loss Landscapes of Machine Learning Models: An Exploratory Study}}

%% authors and affiliations
\author[1]{\normalsize Mingyu Huang}
\author[2]{\normalsize Ke Li}
\affil[1]{\normalsize School of Computer Science and Engineering, UESTC, Chengdu, PR China}
\affil[2]{\normalsize Department of Computer Science, University of Exeter, EX4 4QF, Exeter, UK}
\affil[\Faxmachine\ ]{\normalsize \texttt{m.huang.gla@outlook.com}; \texttt{k.li@exeter.ac.uk}}

\date{}
\maketitle

\vspace{-3ex}
\begin{abstract}
  Previous efforts on hyperparameter optimization (HPO) of machine learning (ML) models predominately focus on algorithmic advances, yet little is known about the topography of the underlying \textit{hyperparameter (HP) loss landscape}, which plays a fundamental role in governing the search process of HPO. While several works have conducted \textit{fitness landscape analysis} (FLA) on various ML systems, they are limited to properties of \textit{isolated} landscape without interrogating the potential \textit{structural similarities} among them. The exploration of such similarities can provide a novel perspective for understanding the mechanism behind modern HPO methods, but has been missing, possibly due to the expensive cost of large-scale landscape construction, and the lack of effective analysis methods. In this paper, we mapped $1,500$ HP loss landscapes of $6$ representative ML models on $63$ datasets across different fidelity levels, with $11$M$+$ configurations. By conducting exploratory analysis on these landscapes with fine-grained visualizations and dedicated FLA metrics, we observed a similar landscape topography across a wide range of models, datasets, and fidelities, and shed light on several central topics in HPO.
\end{abstract}

\section{Introduction}
\label{sec:introduction}
\textbf{Hyperparameter optimization.} In the past decade, considerable efforts have been invested in developing hyperparameter optimization (HPO) techniques to automate the laborious task of hyperparameter (HP) tunning of machine learning (ML) models, and many successful methods (e.g,~\cite{SrinivasKKS10,HutterHL11,BergstraBBK11,SnoekLA12,BergstraB12,KarninKS13,LiJDRT17,FalknerKH18,AwadMH21}, and see surveys~\cite{HKV2019,BischlBLPRCTUBBDL23,KarmakerHSXZV22,YangS20}) have significantly advanced this field. HPO is often cast as a \textit{black-box optimization problem}, where the goal is to \textit{efficiently} search for an HP configuration $\bm{\lambda} \in \bm{\Lambda} = \Lambda_1 \times \dots \times \Lambda_n$ for an ML algorithm $\mathcal{A}$ that yields a loss value $\mathcal{L}(\bm{\lambda})$ as small as possible on a given dataset $\mathcal{D}$. By ``black-box'' it means that while we can evaluate $\mathcal{L}(\bm{\lambda})$ for any $\bm{\lambda} \in \bm{\Lambda}$, we have no access to the explicit form of the loss function $\mathcal{L}: \bm{\Lambda} \to \mathbb{R}$, nor other information like the gradients or the Hessian.

\textbf{The lack of problem understanding.} This lack of knowledge on $\mathcal{L}$ in turn renders HPO solvers to function as black boxes as well, and many of them are often based on rough intuition and assumptions. Albeit their prominent empirical \textit{output performance}, there is a lack of understanding of the \textit{inside mechanism} to support their reliability and validity. Consequently, this poses challenges for pinpointing bottlenecks of the optimizers and thus further improvement. Even worse, such lack of transparency also hinders the development of human trust in HPO, hampering its wide-spread application~\cite{BischlBLPRCTUBBDL23,DrozdalWWDYZMJS20}. Addressing these requires a deep understanding of the global topography of the loss function surface, which is, however, non-trivial due to its black-box nature. Also, as pointed out by a position paper~\cite{HerrmannLECWFRHBB24} at ICML'24 [$\cdots$ in major ML venues, there is little incentive to engage in exploratory ML research].

\begin{wrapfigure}{r}{.3\textwidth}
    \centering
    \includegraphics[width=\linewidth]{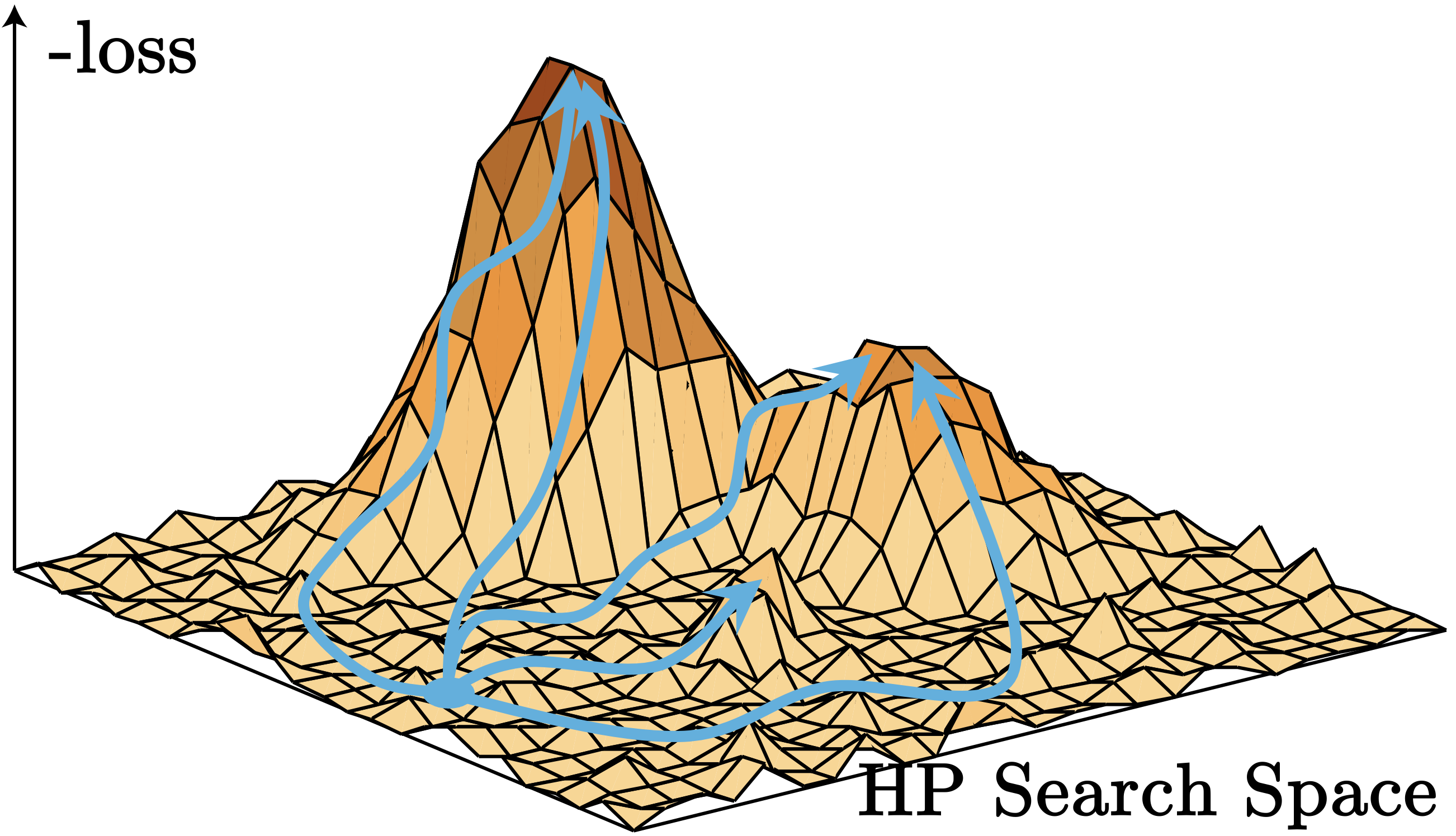}
    \caption{Fitness landscape of an illustrative $2$D problem.}
    \label{fig:demo_landscape}
\end{wrapfigure}

\textbf{Fitness landscape.} The \textit{fitness landscape} metaphor, which was pioneered by Sewall Wright nearly a century ago~\cite{Wright32}, is a fundamental concept in evolutionary biology~\cite{VaishnavBMYFATLCR22,PapkouGEW23,VisserK14,AguilarPW17} and has been widely adapted for understanding black-box systems across various domains including combinatorial optimization \cite{HuangL23, Tayarani-NajaranP14, Prugel-BennettT12} and physical chemistry~\cite{Doye02,WangV03}. Intuitively, it can be conceived as a (hyper-)surface as formed by \textit{fitness} ($\mathcal{L}$ in our context) values across a high-dimensional configuration space (\pref{fig:demo_landscape}). Each spatial location in this landscape represents a configuration (i.e., $\bm{\lambda} \in \bm{\Lambda}$), with its \textit{elevation} indicating the fitness (i.e., $\mathcal{L}(\bm{\lambda})$) and optimal configurations sitting on the valleys. As an optimization process is akin to navigating downhill towards these valleys (e.g., uphill trajectories in~\pref{fig:demo_landscape}), the topography of the underlying landscape contains essentially all the information needed for understanding system properties. Over the past several decades, a plethora of \textit{fitness landscape analysis} (FLA) techniques~\cite{VisserK14,MalanE13,Malan21} have been developed to characterize this topography, e.g., ruggedness~\cite{Weinberger90}, modality~\cite{OchoaTVD08}, neutrality~\cite{StadlerR01}, among many others.

\textbf{Current gaps.} While several works~\cite{PushakH22, TeixeiraP22, PimentaSOP20, SchneiderSPBTK22, MohanBDL23} have explored various properties of the \textit{HP loss landscapes} of ML models, they are often in a \textbf{restricted scope} (e.g., only a handful of landscapes), which limits the generalizability of their findings. More importantly, prior works often focus on \textit{individual} landscapes and fail to interrogate the potential \textbf{structural similarities}~\cite{HuangL23} among \textit{different} landscapes induced on distinct models, datasets, training fidelities, and loss functions, which are concerned by many central topics in HPO, (e.g., transfer learning~\cite{SwerskySA13} and multi-fidelity optimization~\cite{LiL24}). Yet, the exploration of such similarities does not simply equate to an additive extension of prior works, but further requires both \textit{qualitative} and \textit{quantitative} \textbf{FLA methods} that could enable comprehensive \textit{comparisons} among different HP loss landscapes. Unfortunately, there are no bespoke FLA metrics for this purpose, and existing landscape visualization approaches~\cite{Michalak19, BiedenkappMLH18, WalterWC22, Friedman01, AkibaSYOK19} struggle to tackle high-dimensionality while preserving neighborhood structure at the same time.

\textbf{Key considerations.} In this paper, we mapped the largest set of over $\bm{1,500}$ HP loss landscapes induced on $\bm{6}$ representative ML models and $\bm{63}$ datasets, with a total of more than $\bm{11}$\textbf{M} evaluated HP configurations across different levels of fidelities. To enable comprehensive analysis and comparisons among these landscapes, we developed an FLA framework that incorporates: $\blacktriangleright$ a neighborhood-aware HP loss landscape visualization method applicable to high-dimensions, $\blacktriangleright$ $6$ FLA metrics quantifying landscape topologies from complementary perspectives, and $\blacktriangleright$ $3$ similarity metrics that leverage rankings of HP configurations to allow for informative landscape similarity assessment in the HPO context. With these, we shed light on the following HPO scenarios:

\begin{itemize}[leftmargin=2.5ex,topsep=0pt,itemsep=-1ex,partopsep=-0.25ex,parsep=1.5ex]
    \item \textbf{HP loss landscapes on training versus test data.} Overfitting is one of the biggest concerns in ML~\cite{Ng97,CaruanaLG00,RechtRSS19,BelkinHM18,RoelofsSRFHMS19,IshidaYS0S20}. While there has been empirical investigation on adaptive overffiting~\cite{RechtRSS19} with \textit{scattered} data points from Kaggle competitions~\cite{RoelofsSRFHMS19}, there is a lack of more \textit{structured} understanding, e.g., \textit{how the topography of the training loss landscape resemble that of its test counterpart?}    
    \item \textbf{HP loss landscapes with different fidelities.} Given the time-demanding nature of model evaluation, multi-fidelity HPO methods~\cite{KarninKS13, KandasamyDOSP16, LiJDRT17, KandasamyDSP17, FalknerKH18, AwadMH21} have achieved prominent performance by more efficient resource allocation. However, while there exists an implicit assumption: \textit{the topography of HP loss landscapes with lower fidelities would stay close to the groundtruth}~\cite{PushakH22,BischlBLPRCTUBBDL23}, it has only been \textit{indirectly} verified by algorithmic performances, without landscape-level evidence.
    \item \textbf{HP loss landscapes across datasets.} Leveraging priors obtained from previous tasks to expedite the learning process for new tasks is another promising direction for HPO~\cite{SwerskySA13,BardenetBKS13,WistubaSS15b,FeurerSH15,WistubaSS15c,KimKC17,Vanschoren18,RakotoarisonMRS22}. However, despite the prevalence of such transfer learning methods, there is a lack of understanding of how \textit{HP loss landscapes of the same model induced on different datasets would compare to each other?} 
    \item \textbf{HP loss landscapes across models.} While it may seem intuitive that HP loss landscapes would differ depending on the target ML model, in practice the fact is often that common HPO methods could perform robustly under different ML systems. This inspires us to investigate \textit{whether the general family of HP loss landscapes may share certain inherent structural similarities.}
\end{itemize}

\textbf{Contributions.} In a nutshell, our main contributions are in three folds: $\blacktriangleright$ We constructed the largest set of empirical HP loss landscapes in literature. $\blacktriangleright$ We developed a bespoke FLA framework for studying structural similarities across different landscapes, with improved visualization capability and comprehensive metrics. $\blacktriangleright$ We conducted the first exploratory study on the relationships between HP loss landscapes induced on different $i)$ ML models, $ii)$ datasets, $iii)$ fidelities and $iv)$ training versus test sets, showing that the general family of HP loss landscapes could share similar structural characteristics, and providing insights to explain the effectiveness of many prevalent methods in HPO. The FLA framework and collected landscape data will be made publicly available upon acceptance. 

Our value proposition in this paper hinges on the hypothesis that with a deeper understanding of the underlying HP loss landscapes of ML systems, we can advance the interpretability and trustworthiness of the HPO process, and enhance the prospects of it as a tool to benefit broader scientific domains.

\section{HP Loss Landscape Construction and Analysis Method}
\label{sec:methodology}
% \vspace{-1.5em}
This section introduces our FLA framework developed for exploratory analysis of HP loss landscapes. We begin with a formal introduction of the HP loss landscape and how it can be constructed. We then provide a brief overview of our fine-grained landscape visualization method, selected classical FLA features, as well as our quantitative measures for landscape similarity assessment. Due to the stringent page limit, we only provide a sketch of these methods and more details are in~\pref{app: appendix_methods}. 

\textbf{HP loss landscape.} Formally, it can be constructed as a triplet $\langle\bm{\Lambda},\mathcal{L},\mathcal{N}\rangle$: $\blacktriangleright$ a \textbf{search space} $\bm{\Lambda}$ containing the set of sampled (or if possible, enumerated) configurations from the feasible region, $\blacktriangleright$ a \textbf{loss function} $\mathcal{L}$ assigning a loss value for each $\bm{\lambda} \in \bm{\Lambda}$, and $\blacktriangleright$ a \textbf{neighborhood structure} $\mathcal{N}$, which relates configurations in $\bm{\Lambda}$ geographically, i.e., specifies which configurations are neighbors to each other. This in turn depends on a \textit{distance} function $d: \bm{\Lambda} \times \bm{\Lambda} \to \mathbb{R}$, e.g., Hamming distance for categorical HPs, Euclidean or Manhattan distance for numerical HPs. The neighborhood $\mathcal{N}$ can then be defined via a distance threshold $\delta$, where $\bm{\lambda}_j\in\mathcal{N}(\bm{\lambda}_i)$ if $d(\bm{\lambda}_j,\bm{\lambda}_i)\leq\delta$. While this is a rather general formulation, we will introduce our specific choices of these ingredients in the next section. 

\textbf{Landscape construction as directed graphs.} Given the inherent neighborhood structure in HP loss landscapes, representing it as a \textit{graph} is then a natural choice. In this model, each vertex is a configuration $\bm{\lambda} \in \bm{\Lambda}$, with the associated loss $\mathcal{L}(\bm{\lambda})$ as node attribute. Each $\bm{\lambda}^\prime\in\mathcal{N}(\bm{\lambda})$, is connected to $\bm{\lambda}$ via a \textit{directed edge}, where the direction of each edge depends on the relative values of $\mathcal{L}(\bm{\lambda})$ and $\mathcal{L}(\bm{\lambda}^\prime)$, always pointing to the \textit{fitter} one. Such a graph-based model for the HP loss landscape inherently captures its neighborhood structure and loss dynamics. Once constructed, it then allows the implementation of many FLA methods via highly efficient graph traversal algorithms.

\textbf{Landscape topography} plays a crucial role in governing the search behaviors of HPO algorithms. To \textit{quantitatively} characterize its features, we pick up $6$ dedicated FLA metrics, each focusing on a specific aspect, and together they offer a comprehensive view of the essential landscape properties: 

\begin{itemize}[leftmargin=2.5ex,topsep=0pt,itemsep=-1ex,partopsep=-0.5ex,parsep=1.5ex]
    \item \textit{Autocorrelation}~\cite{Weinberger90}: $\rho_a \in [-1, 1]$, it measures the ruggedness of the landscape (i.e., degree of fluctuations in fitness values), where a higher value implies a smoother landscape surface.
    \item \textit{Loss assortativity}, $\mathcal{L}_{\text{ast}} \in [-1, 1]$: adapted from the general assortativity measure of complex networks~\cite{Newman03}, this shows the tendency that configurations with similar loss values are neighbors to each other, where a large value implies more local clustering.
    \item \textit{Mean neutrality}, $\bar{\nu} \in [0, 1]$: it is the Spearman correlation between the configurations neutralities and their distances to the global optimum; it quantifies the proportion of search space where fitness values remain constant (up to a tolerance $\epsilon=0.1\%$ in ML context) in local neighborhoods;
    \item \textit{Neutrality distance correlation}, $\text{NDC} \in [-1, 1]$: it assesses the tendency that the terrain becomes flatter when approaching the global optimum (i.e., diminishing returns).
    \item \textit{$\#$Local optima}, $n_{\text{lo}}$, and \textit{mean basin size}, $\bar{s}_{\mathcal{B}}$: local optima with a large basin of attraction can pose a greater challenge to optimizers, and so does a larger number of local optima.
\end{itemize}

\textbf{Landscape visualization} is crucial for enabling an \textit{intuitive} understanding of a landscape's topography, but is notoriously difficult because it is often in a \textit{high-dimensional} space. Some prior works address this by simply plotting only one or two HP(s) each time~\cite{Friedman01, AkibaSYOK19}, which fail to capture the global topography. Others works directly applied dimensionality reduction techniques to project the landscape into a $2$D space \cite{Michalak19, BiedenkappMLH18, WalterWC22}, but the results are not able to preserve the \textit{local neighborhood} structure. We tackle this by leveraging a graph embedding method, HOPE~\cite{OuCPZ016}, to first learn a vector representation for each configuration in the landscape, which can preserve high-order proximities between nodes. We then compress the obtained embeddings into $2$D using the UMAP~\cite{McInnesH18} algorithm. To further refine interpretability, we additionally apply linear interpolation to generate a continuous landscape surface. We provide more details on design choices and configurations in~\pref{app:app_method_visual}.

\textbf{Landscape similarity assessment.} Though general landscape features or embeddings can be directly used to compare different landscapes~\cite{SchneiderSPBTK22, HuangL23}, additional \textit{domain knowledge} can often lead to more informative results~\cite{Smith-Miles08}. To put this into HPO context, we ground our similarity measures for HP loss landscapes on the consistency of their \textit{performance rankings}, i.e., whether a highly fit configuration on one landscape also tends to excel on another. For this purpose, we use $3$ complementary metrics: $\blacktriangleright$ \textit{Spearman's $\rho_s$}~\cite{Spearman61}, it indicates the general consistency of performance ranks of configurations in two landscapes; $\blacktriangleright$ Kaggle's \textit{Shake-up} metric~\cite{Trotman19}, it assesses the expected movement of configuration rankings across two landscapes; $\blacktriangleright$ the $\gamma$\textit{-set similarity}~\cite{WatanabeAOH23}, it quantifies the ratio of overlaps between the most prominent regions (as defined by $\gamma$) of two landscapes divided by their unions.

\section{Experimental Setup}
\label{sec:setup}

\begin{table}
  \caption{Summary of the models and landscape settings included in our empirical study.}
  \centering
  \fontsize{8pt}{9pt}\selectfont
  \begin{tabular}{
        >{\raggedright\arraybackslash}m{0.075\linewidth}
        >{\raggedright\arraybackslash}m{0.05\linewidth}
        >{\centering\arraybackslash}m{0.05\linewidth}
        >{\raggedleft\arraybackslash}m{0.05\linewidth}
        >{\raggedright\arraybackslash}m{0.315\linewidth}
        >{\raggedleft\arraybackslash}m{0.075\linewidth}
        >{\raggedleft\arraybackslash}m{0.1\linewidth}
      }
      \hline
      \addlinespace[0.05cm]
      \textbf{Group} & \textbf{Model} & \textbf{\#HPs} & \textbf{\#Tasks} & \textbf{\#Fidelities} & \textbf{\#Configs.} & \textbf{\#Landscapes} \\
      \hline
      \addlinespace[0.05cm]
      \multirow{2}{*}{\centering \textbf{Classical}} & \texttt{DT} & $5$ & $57$ &  $\alpha \in \{0.1, 0.25, 1\}$ & $24,200$ & $342$\\
      & \texttt{SVM} & $4$ & $-$ & $\alpha \in \{0.1, 0.25, 1\}$ & $2,340$ & $-$ \\ % Example values for SVM
      \hline
      \addlinespace[0.05cm]
      \multirow{3}{*}{\centering \textbf{Ensemble}} & \texttt{XGB} & $5$ & $57$ & $\alpha \in \{0.1, 0.25, 1\}$ & $14,960$ & $342$ \\
      & \texttt{RF} & $6$ & $57$ & $\alpha \in \{0.1, 0.25, 1\}$ & $11,250$ & $342$ \\
      & \texttt{LGBM} & $5$ & $57$ & $\alpha \in \{0.1, 0.25, 1\}$ & $13,440$ & $342$ \\
      \hline
      \addlinespace[0.05cm]
      \multirow{3}{*}{\centering \textbf{DNNs}} & \texttt{CNN} & $8$ & $6$ & $\alpha \in \{0.1, 0.25, 1\}$, $E \in \{10, 25, 50\}$ & $6,480$ & $108$ \\
      & \texttt{NB101} & NAS & $1$ & $E \in \{4,12, 36, 108\}$ & $432,624$ & $8$ \\
      & \texttt{NB201} & NAS & $3$ & $E \in \{10,50,200\}$ & $15,625$ & $18$\\
      & \texttt{NB360} & NAS & $2$ & $E \in \{10,50,200\}$ & $15,625$ & $12$\\
      \hline
  \end{tabular}
  \label{tab:setup}
\end{table}

Investigating HPO scenarios posed in~\pref{sec:introduction} demands the construction of HP loss landscapes on a wide spectrum of models, tasks and fidelities, and a vast amount of HP configurations. To this end, we conduct large-scale experiments to collect such data, as summarized in~\pref{tab:setup}, which consumes over $7$ thousand of GPU and CPU hours (\pref{app:resources}) and ends up in $1,500+$ distinct landscapes.

\textbf{Models.} To ensure its generalizability, we study $3$ mainstream ML models for our empirical study: $\blacktriangleright$ We start with $2$ classical models, decision tree (\texttt{DT}) and support vector machine (\texttt{SVM}). $\blacktriangleright$ Then, we study $3$ types of tree ensembles: random forest (\texttt{RF}), \texttt{XGBoost}~\cite{ChenG16} and \texttt{LightGBM}~\cite{KeMFWCMYL17}, which remain the most competitive~\cite{GrinsztajnOV22} on tabular tasks. $\blacktriangleright$ Finally, we investigate deep neural networks (DNNs), convolutional neural networks (\texttt{CNN})~\cite{KrizhevskySH12} in particular, using a joint architecture and HP search space~\cite{BansalSJZH22}. To allow the investigation of more diverse CNN architectures, we additionally explore $3$ popular neural architecture search (NAS) benchmarks. These include NASBench-101~\cite{YingKCR0H19}, NASBench-201~\cite{Dong020}, and NASBench-360~\cite{TuRKSST22}, which offer evaluations of tens of neural architectures under different tasks, and fidelities (see~\pref{app:nasbenchs}).

\textbf{HP search spaces.} Due to the presence of continuous HPs, a full enumeration of many HP spaces is not possible. Instead, following~\cite{PushakH22}, we adopt a grid sampling approach to construct HP loss landscapes. Note that while grid search is less \textit{efficient} than methods like random search in terms of optimization~\cite{BergstraB12}, it can yield more \textit{interpretable} landscapes (e.g., the neighboring configurations only differ in one HP) that are favorable by FLA. Also, though landscape extrapolation via surrogate models can offer infinite sampling resolution~\cite{VaishnavBMYFATLCR22,MohanBDL23}, they introduce additional biases to the analysis~\cite{FahlbergFHR23}. An important consideration in designing the grid spaces for each model is to ensure that they are representative of the settings commonly used in real-world practice, while keeping a good balance between grid resolution and computational cost. In doing so, we strictly followed our devised protocols in~\pref{app:appendix_search_space} to create each space, with typically $2$ to $8$ HPs, and a few to tens of thousands of configurations (\pref{tab:setup}; \pref{app:appendix_search_space}). These spaces are not meant to cover the entire feasible region, but rather to focus on more prominent and informative regions as in HPO. 

\textbf{Datasets and configuration evaluation.} For the first two groups of models, except \texttt{SVM}, we ground our analysis on $57$ tasks (\pref{app: appendix_datasets}) from the tabular benchmark proposed in~\cite{GrinsztajnOV22}, which span a broad range of complexities (i.e., number of entries and features). For \texttt{SVM}, due to its poor scalability, we are only able to evaluate it on a few tasks. As for \texttt{CNN}, we consider $6$ image classification tasks as detailed in~\pref{app: appendix_datasets}. For each task, unless predefined, we randomly split the data into training ($80\%$) and test ($20\%$) sets. For all HP configurations $\bm{\lambda}\in\bm{\Lambda}$ of each model, we exhaustively evaluate $\mathcal{L}(\bm{\lambda})_{\text{train}}$ and $\mathcal{L}(\bm{\lambda})_{\text{test}}$ using $5$-fold cross-validation. We use root mean squared error (RMSE) and $R^2$ score to serve as the loss function $\mathcal{L}$ for regression tasks, while for classification, we use accuracy and ROC-AUC. We control the fidelity of the training by either varying the ratio of training data ($\alpha$) or the number of training epochs ($E$,~\pref{tab:setup}), thus obtaining cheaper proxies of the loss, $\mathcal{L}_{\text{LF}}$.

\textbf{Distances and neighborhood.} After obtaining the set of HP configurations $\bm{\Lambda}$ and their corresponding losses $\mathcal{L}$ on each scenario, we then follow \cite{PushakH22} to assess the distance between \textit{categorical} HPs with the Hamming distance, where each differing value is assigned a $d=1$; for \textit{numerical} HPs, we use the Manhattan distance, to reflect the number of discrete steps between them on the search grid. The aggregate distance between two configurations, $\bm{\lambda}_i$ and $\bm{\lambda}_j$, is the sum of these component distances, and we say they are neighbors to each other if $d(\bm{\lambda}_j,\bm{\lambda}_i)=1$. Such a neighborhood structure imitates the manual ``baby-sitting'' tuning process~\cite{YangS20} and preserves the local proximity of configurations.

\section{Results and Analysis}
\label{sec:results}
We now investigate the obtained HP loss landscapes under the $4$ scenarios posed in~\pref{sec:introduction}. We begin with exploring the general characteristics of these landscapes in \pref{sec: results_1}. We then compare landscapes under: $i)$ training and test setups (\pref{sec: results_2}), $ii)$ different fidelities (\pref{sec: results_3}), and $iii)$ different datasets (\pref{sec: results_4}). Note that as we only constructed a few landscapes for \texttt{SVM}, we left its results in~\pref{app:svm}, where our findings here also generalize to \texttt{SVM}.

\begin{figure*}[t!]
    \centering
    \includegraphics[width=\linewidth]{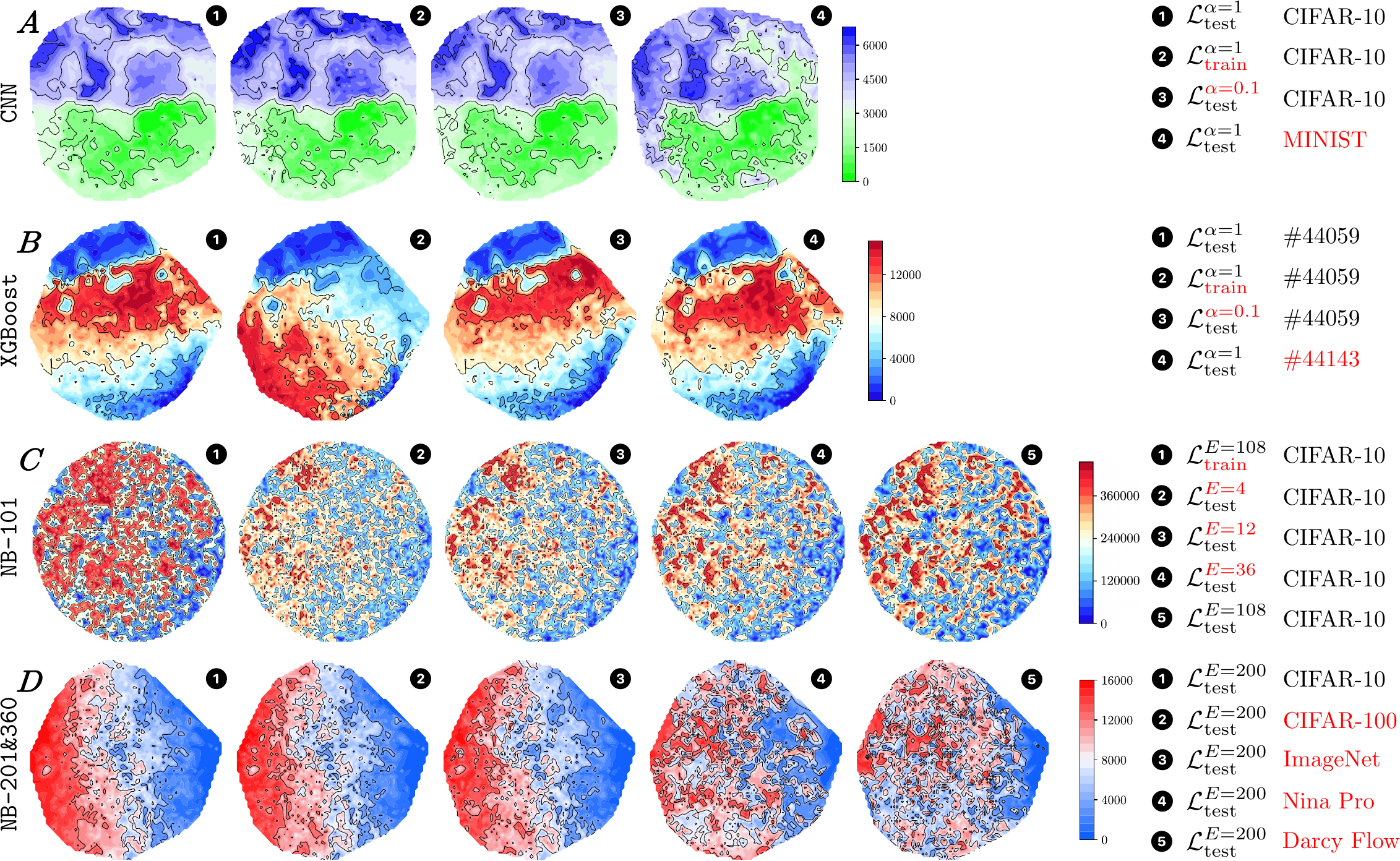}
    \caption{$2$D visualization of HP loss landscapes using our proposed method in~\pref{sec:methodology} for \textbf{(A)} \texttt{CNN} and \textbf{(B)} \texttt{XGBoost}, w.r.t. $\mathcal{L}_{\mathrm{test}}$ and $\mathcal{L}_{\mathrm{train}}$ as well as different datasets and fidelities (by varying the fraction of training data $\alpha$ or the number of training epochs $E$, marked in \textcolor{red}{red color}). \textbf{(C)} shows the visualization of the NASBench-101 landscape with different training epochs on CIFAR-10, and \textbf{(D)} visualizes the NASBench-201 (\ding{182}--\ding{184}) and NASBench-360 (\ding{185}--\ding{186}) loss landscapes of different datasets. Colors indicate ranks of performance (lower rank values are better) to normalize the losses.}
    \label{fig:landscapes_all}
\end{figure*}

\begin{figure*}[t!]
    \centering
    \includegraphics[width=\linewidth]{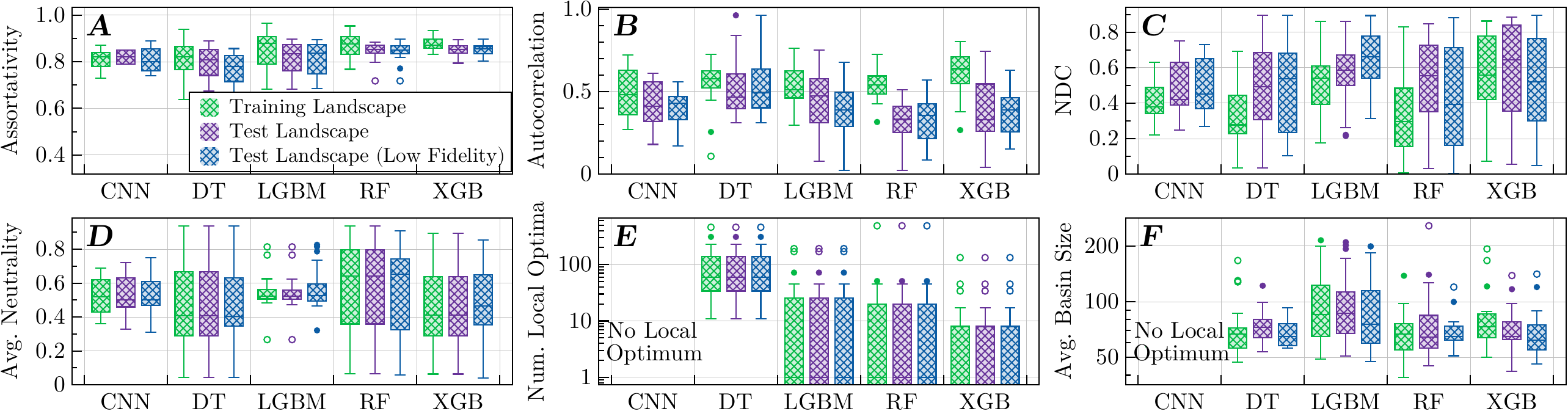}
    \caption{Distribution of FLA metrics across all datasets for landscapes of $1)$ $\mathcal{L}_\mathrm{test}$, $2)$ $\mathcal{L}_\mathrm{train}$ and $3)$ $\mathcal{L}_{\mathrm{test}}$ obtained with $10\%$ training data.}
    \label{fig:landscape_metrics}
\end{figure*}

\subsection{Overall Characteristics of Studied HP Loss Landscapes}
\label{sec: results_1}

We first visualize the topography of the $\mathcal{L}_{\mathrm{test}}$ landscapes for each model using the method in~\pref{sec:methodology}. \pref{fig:landscapes_all} (A-1) and (B-1) show the results for \texttt{CNN} and \texttt{XGBoost}. These provide a general impression of these landscapes, and we clearly observed certain \textit{structural similarities}: \textit{the HP loss landscapes of the studied models are \textcolor{brightmaroon}{relatively smooth}; configurations are \textcolor{brightmaroon}{locally clustered} in terms of their losses; there is a \textcolor{brightmaroon}{highly distinguishable plateau} consisting of prominent configurations, where the terrain becomes \textcolor{brightmaroon}{flatter}.} As depicted in~\pref{app:addi_visual}, the landscapes of the other models also exhibit similar patterns. We will then take a closer look at these patterns with quantitative FLA metrics in~\pref{fig:landscape_metrics}. 

\textbf{Remark 1.} \textit{We further verified the reliability of these visualizations in~\pref{app:nk} with tunable $NK$-landscapes~\cite{kauffman93} and the black-box optimization benchmark functions from \texttt{COCO} platform~\cite{HansenARMTB21}.}

\begin{figure*}[t!]
    \centering
    \includegraphics[width=\linewidth]{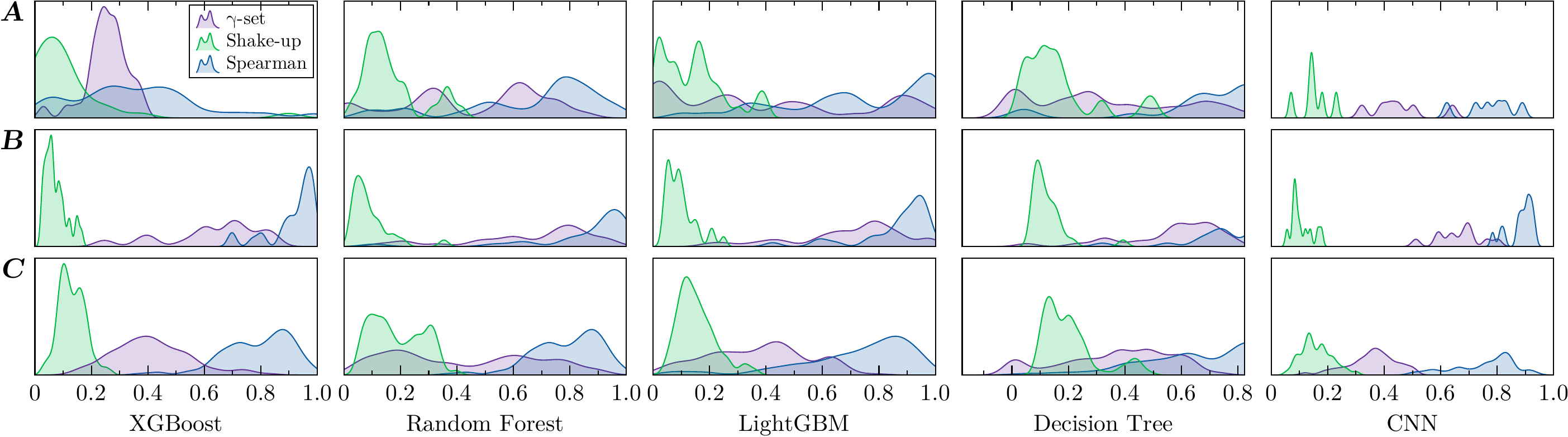}
    \caption{Distribution of Spearman, Shake-up and $\gamma$-set metrics between \textbf{(A)} $\mathcal{L}_\mathrm{test}$ and $\mathcal{L}_\mathrm{train}$, \textbf{(B)} $\mathcal{L}_\mathrm{test}$ and $\mathcal{L}_{\mathrm{test}}$ with $10\%$ training data, \textbf{(C)} $\mathcal{L}_\mathrm{test}$ across datasets.}
    \label{fig:landscape_sim}
\end{figure*}

\textbf{Remark 2.} \textit{The patterns observed for the systems in~\pref{fig:landscapes_all} are representative of the other systems, as justified quantitatively by the landscape metrics in~\pref{fig:landscape_metrics} and landscape similarities in~\pref{fig:landscape_sim}.}

\textbf{Fairly smooth and locally clustered.} From~\pref{fig:landscape_metrics}, we found high loss assortativity and autocorrelation values for the majority of our studied $\mathcal{L}_{\mathrm{test}}$ landscapes. These imply that configurations with similar $\mathcal{L}_{\mathrm{test}}(\bm{\lambda})$ tend to be neighbors to each other, where a small change in $\bm{\lambda}$ is not likely to cause drastic variation in $\mathcal{L}_{\mathrm{test}}(\bm{\lambda})$ (i.e., a fairly smooth landscape). Such property is highly beneficial, as it makes an elegant separation of the prominent regions from the poorly performing ones possible. Such separation can be known as \textit{priors}, and be utilized to guide the design of search spaces that focus on the promising region~\cite{LiSJBZZC22}, or to warm-start HPO solvers directly within that region~\cite{FeurerSH15}. Alternatively, this knowledge can also be \textit{learned on-the-fly} as the optimization proceeds, and thus guides the solver towards fitter areas~\cite{SnoekLA12}, or assisted in progressively pruning the search space to narrow down the search~\cite{PerroneS19,WangFT20}. On the contrary, if the landscape is highly rugged instead, in which prominent configurations are often \textit{interspersed} with poorer ones, it would become more difficult to distinguish a clear advantageous region, making it more difficult to navigate the landscape or design a compact search space. 

\textbf{Nearly unimodal.} As shown in~\pref{fig:landscape_metrics} (E), we found a considerable fraction of the $\mathcal{L}_{\mathrm{test}}$ landscapes are \textit{unimodal}, while for some landscapes (\texttt{DT}, especially), there could be a handful to dozens of local optima. This is somewhat contradictory to~\cite{PushakH22} at first glance, in which it is found that almost all landscapes studied are nearly unimodal. However, when taking a closer look at the local optima in our landscapes, we found that they are usually featured in a small basin of attraction (see \pref{fig:landscape_metrics} (F)). This makes them relatively \lq\lq shallow\rq\rq, and would not pose significant obstacles to optimization. Thus, we contend that the unimodal hypothesis is still valid for most of our studied scenarios.

\textbf{Highly neutral; planar around the optimum.} As in~\pref{fig:landscape_metrics} (D), we clearly observed that HP loss landscapes are often featured in high neutrality. This indicates that a large portion of purturbations in the landscapes will result in subtle changes in $\mathcal{L}_{\mathrm{test}}$ (i.e., $\leq 1\text{\textperthousand}$). We postulate that a major reason for this is the \textit{low effective dimensionality}~\cite{BergstraB12} of HPO problems, i.e., usually only a small subset of all tunable HPs have a significant impact on performance. As our designed search spaces have excluded HPs that have no obvious impact on performance, the high neutralities of the resulting landscapes imply that neutral plateaus can actually be more prevalent than one may expect. This phenomenon is more pronounced for the well-performing regions, as illustrated by the high NDC values in~\pref{fig:landscape_metrics} (C). It suggests that as we move closer to the global optimum, it is more likely to encounter neutral moves, and the landscape thus becomes flatter. This observation coincides with experimental observations in HPO, i.e., the gain of tuning HPs often progressively diminishes as approaching higher performance. Such property can pose challenges to the solvers, as there is little gradient information that can be utilized for navigating towards fitter configurations, and they can \lq get lost\rq\ in neutral plateaus. This then underscores the necessity of careful search space design prior to optimization.

\textbf{Remark 3.} \textit{The exact measured neutrality of a landscape can depend on multiple factors, including the choice of HPs, the domain, and resolution of the search space, as well as the tolerance $\epsilon$ we used to define neutral moves. Though we have made careful considerations in ensuring the generalizability of our results (see~\pref{app: appendix_setup}), there are no golden standards or consensus for us to explicitly follow. Our goal here is to evoke attention to the potential redundancy in search spaces used in HPO practice, rather than obtaining an absolute measure of neutrality in these landscapes.}

The NAS landscapes, as evidenced in~\pref{app:nas}, also yield high assortativity and autocorrelation values, and the terrain tends to be flatter near the optimum as well. These properties are not surprising as both NAS and HPO are sibling subjects under the umbrella of AutoML. Albeit this, we found that NAS landscapes exhibit considerably more local optima, constituting a multimodal topography (see, e.g., \pref{fig:landscapes_all} C-5). In this regard, the NAS problems seem to be more complex. Therefore, we can expect that directly applying HPO solvers to NAS may not lead to satisfactory performance. Instead, tailored search strategies are required to tackle such more rugged landscape surfaces.

\subsection{HP Loss Landscapes on Training and Test Data}
\label{sec: results_2}

\pref{fig:landscapes_all} (A-2), (B-2) visualize the $\mathcal{L}_\mathrm{train}$ landscapes of \texttt{CNN} and \texttt{XGBoost} (see also~\pref{app:addi_visual}). The \textit{structural characteristics} of these landscapes show high concordance with our previously discussed properties for $\mathcal{L}_\mathrm{test}$. On the other hand, for \textit{performance rankings}, we observed that the $\mathcal{L}_\mathrm{train}$ generally correlate with $\mathcal{L}_\mathrm{test}$ for \texttt{CNN}, \texttt{RF}, \texttt{DT}, and \texttt{LGBM}, whereas for \texttt{XGBoost}, there is significant discrepancy between the landscapes of $\mathcal{L}_\mathrm{train}$ and $\mathcal{L}_\mathrm{test}$. We further quantitatively discuss such observations in the following paragraphs.

\textbf{Landscape characteristics.} From~\pref{fig:landscape_metrics}, we can clearly see that the structural characteristics for $\mathcal{L}_\mathrm{train}$ and $\mathcal{L}_\mathrm{test}$ landscapes are highly consistent for most studied scenarios. We found that $\mathcal{L}_\mathrm{train}$ landscapes tend to yield relatively higher performance assortativity and autocorrelation values and thus a smoother and more structured terrain. Meanwhile, the NDC values are lower in the training scenarios. These observations imply that the $\mathcal{L}_\mathrm{train}$ landscapes are even more benign than $\mathcal{L}_\mathrm{test}$ ones. In addition, we found that neutrality, \#local optima, and basin sizes rarely change between $\mathcal{L}_\mathrm{train}$ and $\mathcal{L}_\mathrm{test}$ landscapes. Notably, \#local optima found in $\mathcal{L}_\mathrm{train}$ and $\mathcal{L}_\mathrm{test}$ landscapes are almost (if not all) identical. This suggests that the \textit{relative} performance in local neighborhoods tends to be retained in the two cases, despite variations in their numerical values and global rankings. 

\textbf{Landscape similarity.} We quantified the similarity between all pairs of $\mathcal{L}_\mathrm{train}$ and $\mathcal{L}_\mathrm{test}$ landscapes using the $3$ metrics introduced in~\pref{sec:methodology} as shown in~\pref{fig:landscape_sim} (A). Overall, we observed that $R(\mathcal{L}_\mathrm{train})$ and $R(\mathcal{L}_\mathrm{test})$ are globally correlated for all models except \texttt{XGBoost}, as indicated by the significant $\rho_s$ values (median $>0.7$) and low Shake-up metrics (median $<0.15$). However, when zooming into the top-$10\%$ regions, we saw that the majority of our studied scenarios reveal low $\gamma$-set similarities. It indicates that the generalization gap is larger in prominent regions where configurations are highly adapted to the training set. This is more severe for \texttt{XGBoost}, where the median $\gamma$-set similarity is only $0.07$, and there is also a low $\rho_s$ value (median = $0.34$) and high Shake-up score (median = $0.25$).

\begin{figure*}[t!]
    \centering
    \includegraphics[width=\linewidth]{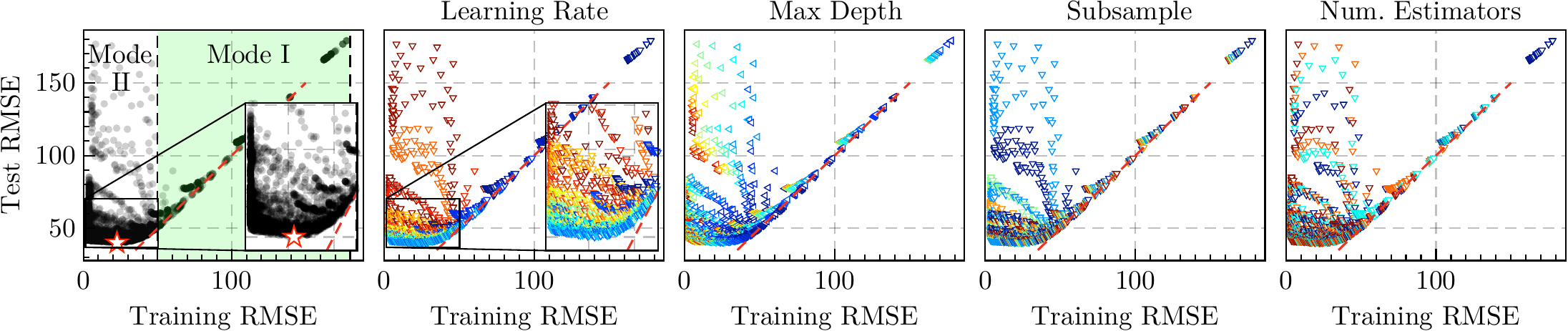}
    \caption{\textbf{(A)} Scatter plot of $\mathcal{L}_\mathrm{train}$ versus $\mathcal{L}_\mathrm{test}$ of \texttt{XGBoost} on the dataset \#44059. $\bm{\lambda}^*_{\mathrm{test}}$ is marked by \textcolor{red}{\ding{73}}, \textbf{(B-E)} The same plot with colors to indicate HPs values. Warmer color indicate higher values.}
    \label{fig:rq1_overfit}
\end{figure*}

\textbf{A closer look at overfitting.} In order to gain more insight into such generalization gaps for \texttt{XGBoost}, we create scatter plots of $\mathcal{L}_\mathrm{test}$ versus $\mathcal{L}_\mathrm{train}$ on the dataset \#44059 in~\pref{fig:rq1_overfit} (A), where we can decompose the pattern into two modes. In the first one, $\mathcal{L}_\mathrm{test}$ highly correlates with $\mathcal{L}_\mathrm{train}$ as it decreases, and the models in this stage \textit{underfit} the data. In the second mode, on the contrary, as points struggle to further move on the $x$-axis ($\mathcal{L}_\mathrm{train}$), they stagnate or even significantly increase on the $y$-axis ($\mathcal{L}_\mathrm{test}$), indicating strong evidence of overfitting. In particular, we can see a plateauing trend near the $x$-axis, where some models overly excel on the training data, but perform poorly on the test set. To further investigate which kinds of configurations are likely to lead to overfitting, we color the points with respect to their HP values as shown in~\pref{fig:rq1_overfit} (B-E). It is interesting to see that the generated plots demonstrate clear patterns between the value of each HP and the resulting performance. In particular, we find that learning rate, max depth, and subsample have a significant impact on $\Delta \mathcal{L}$. However, the generalizability of a learner is not monopolized by a single one of them; instead, it depends on their cumulative interactions. For example, the largest $\Delta \mathcal{L}$s are observed for learners that feature a large learning rate, and deep base trees, combined with a low subsample rate, but any of these HP settings alone does not necessarily lead to the worst case performance. In addition to this, we noticed that such a generalization gap is also correlated some dataset properties such as number of samples or features, and we discuss more about this in~\pref{app:dataset_size}.

\subsection{HP Loss Landscapes with Different Fidelities}
\label{sec: results_3}

\pref{fig:landscapes_all} (A-3) and (B-3) show the low-fidelity loss landscapes (denoted as $\mathcal{L}_{\mathrm{test}}^\mathrm{LF}$) for \texttt{CNN} and \texttt{XGBoost} with $\alpha=0.1$ training data. From the plots, we observed that $\mathcal{L}_{\mathrm{test}}^\mathrm{LF}$ landscapes are highly consistent with $\mathcal{L}_{\mathrm{test}}$ in terms of both structural characteristics and performance rankings. More specifically, as reflected in~\pref{fig:landscape_metrics}, all measured FLA metrics of $\mathcal{L}_{\mathrm{test}}^\mathrm{LF}$ landscapes exhibited little difference compared to $\mathcal{L}_{\mathrm{test}}$ landscapes across most studied scenarios. For performance rankings, \pref{fig:landscape_sim} (B) depicts the distribution of the $3$ similarity indicators between $\mathcal{L}_{\mathrm{test}}^\mathrm{LF}$ and $\mathcal{L}_{\mathrm{test}}$ across all datasets for each model. We observed a high Spearman correlation (median $> 0.85$) between $\mathcal{L}_{\mathrm{test}}$ and $\mathcal{L}_{\mathrm{test}}^\mathrm{LF}$ for most cases, and the $\gamma$-set similarities between the top-$10\%$ configurations are also prominent, with medians larger than $60\%$. These imply that $R(\mathcal{L}_{\mathrm{test}})$ and $R(\mathcal{L}_{\mathrm{test}}^\mathrm{LF})$ are highly consistent for the majority of our studied scenarios and there is a large overlap between the promising regions of the two landscapes. In addition, the Shake-up scores yield low values ($\text{median} < 0.1$), suggesting that on average the difference between $R(\mathcal{L}_{\mathrm{test}})$ and $R(\mathcal{L}_{\mathrm{test}}^\mathrm{LF})$ is less than $10\%$. Note that albeit our general observations, the similarity distributions in \pref{fig:landscape_sim} (B) typically have long tails, and there are cases for which the $\mathcal{L}_{\mathrm{test}}^\mathrm{LF}$ landscapes largely deviate from the ground truth. 

\begin{wrapfigure}{r}{.3\textwidth}
    \vspace{-1em}
    \centering
    \includegraphics[width=\linewidth]{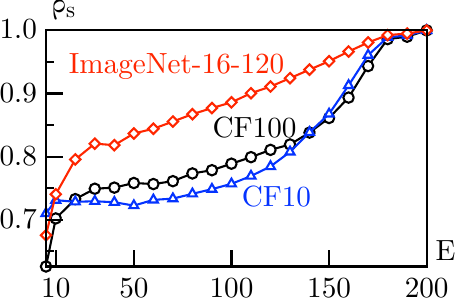}
    \caption{$\rho_s$ between $\mathcal{L}_{\mathrm{test}}^{\textrm{LF}}$ and $\mathcal{L}_{\mathrm{test}}$ in NASBench-201 as a function of training epoch.}
    \label{fig:fidelity}
\end{wrapfigure}

The NAS landscapes also suggest similar findings. In \pref{fig:landscapes_all} (C), we plot the NASBench-101 $\mathcal{L}_{\mathrm{test}}^\mathrm{LF}$ landscapes under four different training epochs on CIFAR-10. While it is not surprising that with higher fidelities (i.e., $E$), the $\mathcal{L}_{\mathrm{test}}^\mathrm{LF}$ landscapes can better approximate the ground truth $\mathcal{L}_{\mathrm{test}}$ landscape, we noticed that with only $4$ training epochs, the $\mathcal{L}_{\mathrm{test}}^\mathrm{LF}$ landscapes can already capture certain prominent regions. We further explored this with NASBench-201, which provides evaluation information after every training epoch. \pref{fig:fidelity} plots the Spearman correlation between $\mathcal{L}_{\mathrm{test}}^\mathrm{LF}$ and $\mathcal{L}_{\mathrm{test}}$ as a function of training epochs $E$. From this we found that with $10$ out of $200$ training epochs, we can already obtain a reasonable ranking.

% Additional results in~\pref{app:nas} demonstrate that the observations on NAS problems are consistent with our findings here. However, note that these findings are based on the statistical analysis regarding the machine learning tasks studied within this paper, while we are aware of some exceptions that are somehow against this conclusion (e.g., the outliers in~\pref{fig:landscape_sim} (B)). Understanding \textit{when} and \textit{why} lower fidelity landscapes can largely deviate from the original one is important for the robustness and accountability of multi-fidelity HPO. Since this paper is concerned with establishing general conclusions, we leave this to future work.

\subsection{HP Loss Landscapes Across Datasets}
\label{sec: results_4}

\begin{wrapfigure}{r}{.55\textwidth}
    \centering
    \includegraphics[width=\linewidth]{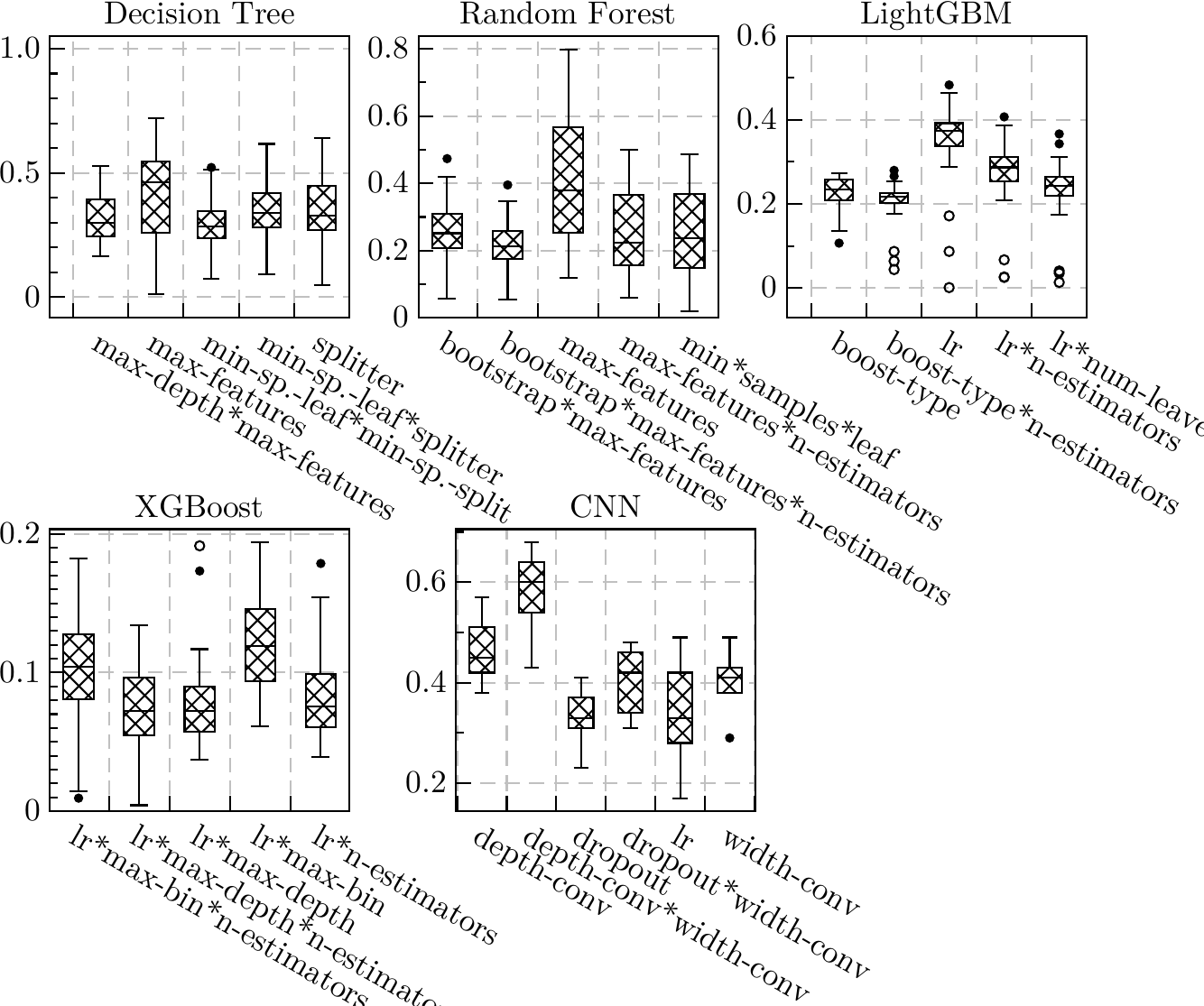}
    \caption{Importance of top-$5$ HPs (or interactions) as indicated by functional ANOVA method of each model.}
    \label{fig:hp_importance}
\end{wrapfigure}

\textbf{Performance rankings.} \pref{fig:landscapes_all} (A-4) and (B-4) show the $\mathcal{L}_{\mathrm{test}}$ landscapes for \texttt{CNN} and \texttt{XGBoost} on different datasets. We can see that the high-level topography of the $\mathcal{L}_{\mathrm{test}}$ landscape are preserved when transferring to a new task. It is interesting to note that the top regions in the original landscape generally retain their positions, despite the changes in their exact contours. The FLA metrics reported in~\pref{fig:landscape_metrics} also support such observation. In addition, from the similarity metrics reported in~\pref{fig:landscape_sim}, we observed that the measured performance reveal clear Spearman correlations (median $>0.65$) across datasets. In practice, stakeholders are more concerned about the top performing regions in the landscape. For this, the overlap between the top-$10\%$ regions, as indicated by the $\gamma$-set similarity, also achieves medians around $40\%$. We also found that the landscape similarities for tabular tasks are correlated with simple dataset meta-features (see \pref{fig:dataset_sim} in \pref{app:dataset_size}), and there have been works that study such features in more depth (e.g., \cite{JomaaSG21}).

\textbf{HP importance and interaction.} In addition to performance rankings, \pref{fig:hp_importance} illustrates the contribution of each HP and their interactions to model performance assessed by the functional ANOVA method. The results indicate that some (or combination terms of) HPs are typically important for many datasets for a given model. For example, the \texttt{depth-conv} contributes a large portion of variance to model performance for \texttt{CNN}, and so does its interactions with \texttt{width-conv}. Such knowledge can also be utilized to expedite the HPO process.

Our additional results for NAS-Bench-201 across CIFAR-10/100 as well as ImageNet-16-120 datasets also revealed similar highly-transferable conclusions (see \pref{fig:landscapes_all} D1-D3). However, again, we note that our findings are based on general trends, and there are outliers in~\pref{fig:landscape_sim} that reveal little landscape similarity between tasks. Also, the results on NASBench-360 (see~\pref{fig:landscapes_all} D4-D5), which evaluates the search space of NASBench-201 on two tasks very different from classic image classification datasets, exhibited considerable deviations from the landscapes in \pref{fig:landscapes_all} D1-D3. Considering that these two tasks are rooted in biomechanics and computational physics respectively, the resulting difference might be reasonable. However, this evokes the need for including a more diverse set of tasks to further validate our findings, and we leave this for future works.

\section{Discussions and Conclusions}
\label{sec:conclusions}

Many studies have significantly advanced the methodological developments of HPO. Yet, the understanding of the underlying HP loss landscapes of ML models as well as their correlations, which are the foundations of HPO, has been largely overlooked. By analyzing $1,500+$ HP loss landscapes of $6$ representative ML models induced on different scenarios with fine-grained visualizations and dedicated FLA metrics, we found a considerable resemblance between test loss landscapes and their training counterparts, albeit the choice of certain HP combinations and datasets can lead to significant discrepancies. We also showed that the topographies of HP loss landscapes with lower fidelities, in general, are highly consistent with the full-fidelity landscapes, and the same applies to landscapes of the same model induced on different datasets. We further revealed a unified portrait of the HP loss landscapes across our studied models in terms of their structural characteristics. For example, they tend to be fairly smooth, nearly unimodal, and featured with a high degree of neutrality.

Our results in part reaffirm isolated patterns observed in prior works~\cite{PushakH22, TeixeiraP22, PimentaSOP20, SchneiderSPBTK22, MohanBDL23} regarding landscape modality and neutrality, and extend them to broader scenarios. At the same time, we provide the first landscape-level evidence to support the effectiveness of transfer learning and multi-fidelity methods, which previously have been mainly based on rough intuition and empirical performance. Moreover, the high-level structural similarities shared by our studied HP loss landscapes induced on different models assist in explaining why many HPO solvers often work robustly across different systems. Also, given the smooth topography of these landscapes, we also highlight the importance of elaborated search space design, including both the selection of HPs and the choice of value domains.

There are three main limitations within this study. $\blacktriangleright$ Despite the data collection effort with more than $7$ thousands of GPU and CPU hours (see \pref{app:resources}), the ML models considered here still do not cover the full spectrum of ML community, particularly more complex DNNs like the Transformers~\cite{VaswaniSPUJGKP17}, GANs~\cite{GoodfellowPMXWOCB14}, and even large language models~\cite{OpenAI23}. While this may disappoint some readers who are more interested in them, such analysis is currently hindered by the overwhelming computational cost required, and we do not intend to draw general conclusions on all ML models here. $\blacktriangleright$ Our landscape construction is based on discrete grid points, which may potentially miss local optima sitting between them. Yet, given the smoothness of the studied landscapes and the relatively small basins of the perceived local optima, such intermediate ones are unlikely to surge in performance, and their basins would be confined to the vicinity of its closest grid points. These are not expected to invalidate our ``nearly unimodal'' finding. $\blacktriangleright$ Our FLA framework currently requires the evaluation of a tremendous number of model configurations, which is extremely expensive. However, we note that as such an exploratory study aims at gathering insights into the underlying problem, it is by nature less time-sensitive. Still, it can be a prominent direction to integrate this analysis framework with existing AutoML platforms like Optuna~\cite{AkibaSYOK19}, or to develop more accurate surrogate models that are able to better preserve landscape topography, particularly for landscapes with sharp peaks (e.g.,~\cite{FahlbergFHR23}).

Despite these caveats, our exploratory study may provide a useful perspective to rethink HPO algorithms by considering the topography of the underlying HP loss landscapes upon which they traverse. There could also be nuanced scrutiny of the scenarios discussed in this study, e.g., the relationship between dataset meta-features and the landscape characteristics. Also, our developed FLA framework could serve as a useful lens for future studies on a broader range of search landscapes in the ML community (e.g., NAS, AutoRL, and AutoML), and beyond (e.g., protein fitness landscapes~\cite{?} toward AI for science). We hope our analysis opens a new avenue towards understanding the fitness landscapes of black-box systems and explainable AI in optimization.

% While numerous efforts have been devoted to developing advanced searching mechanisms and evaluation strategies for HPO, the exploration of the underlying HP search spaces has been largely overlooked in our community. By conducting large-scale exploratory analysis on $1,500$ HP landscapes of $5$ ML models with $>$11M model configurations under different fidelities, we reveal an unified portrait of their topographies in terms of smoothness, neutrality and modality. We also show that these properties are highly transferable across datasets and fidelities, and thus provide fundamental evidence to support the effectiveness of transfer learning and multi-fidelity methods, which previously are mainly based on intuitions. Our developed FLA framework has greatly advanced FLA w.r.t. landscape visualization and similarity quantification, and lays the foundation for more detailed scrutiny at landscapes of a wider range of AutoML problems. We hope this work can inspire more follow-up studies to further explore AutoML landscapes and thus provide more insights to guide the design of new HPO methods.

\begingroup
\small
\bibliographystyle{IEEEtran}
\bibliography{icml2024}
\endgroup

\newpage
\appendix
\section{Summary of Computational Resrouces Used}
\label{app:resources}

All the performance evaluation experiments for \texttt{DT}, \texttt{RF} and \texttt{LGBM} were run on 3 nodes of a cluster, where each node is equiped with Intel$^{\text{\textregistered}}$ Xeon$^{\text{\textregistered}}$ Platinum 8260 CPU@2.40GHz and 256GB memory. As for \texttt{XGB} and \texttt{CNN}, which were run on GPUs, we evaluated the configurations using 8 Nvidia GeForce RTX$^{\text{TM}}$ 3090 GPUs and 8 Nvidia GeForce RTX$^{\text{TM}}$ 2080Ti GPUs. These in total consumed $7$ thousand of CPU and GPU hours to collect all the performance data.

For the landscape construction and analyses, all the experiments were carried out on a single node with Intel$^{\text{\textregistered}}$ Core$^{\text{TM}}$ i9-12900k CPU@5.2GHz and 64GB memory.

\section{Details of the FLA Framework}
\label{app: appendix_methods}

\subsection{Landscape Visualization Method}
\label{app:app_method_visual}

\textbf{HOPE node embedding.} To preserve the intrinsic neighborhood relationship of HP configurations, our proposed landscape visualization method first needs to learn a low-dimensional embedding for each configuration in the landscape (i.e., each node in the graph). In this paper, we choose HOPE\footnote{We use the implementation in \href{https://github.com/benedekrozemberczki/karateclub}{Karateclub} package.} node embedding method to serve this purpose, as it is able to capture asymmetric high-order proximity in directed networks. Specifically, for a directed network\footnote{Most manipulations and analyses of HP loss landscapes as complex networks are grounded on \href{https://github.com/networkx/networkx}{NetworkX} and \href{https://pandas.pydata.org/}{Pandas} package.}, if there is a directed edge from vertex $v_i$ to vertex $v_j$ and from vertex $v_j$ to vertex $v_k$, it is more likely to have a edge from $v_i$ to $v_k$, but not from $v_k$ to $v_i$. In order to preserve such asymmetric transitivity, HOPE learns two vertex embedding vectors $U^s,U^t\in\mathbf{R}^{|V|\times d}$, which is called source and target embedding vectors, respectively. After constructing the high-order proximity matrix $S$ from $4$ proximity measures, i.e., Katz Index, Rooted PageRank, Common Neighbors and AdamicAdar. HOPE finally learns vertex embeddings by solving the a matrix factorization problem: 
\begin{equation}
  \min_{U_s,U_t}\|S-{U^s U^{t^T}}\|_F^2
\end{equation}
\textbf{UMAP dimensionality reduction.} While HOPE (as well as any other node embedding methods) could generate vectorized embeddings for configurations, these embeddings typically are still high-dimensional and not compatible for $2$D visualization. In light of this, we further apply UMAP\footnote{We use the implementation in~\href{https://umap-learn.readthedocs.io/en/latest/}{UMAP} package.} to project the HOPE embeddings into a $2$D space, which is based on three assumptions, and here we provide brief discussion on whether they hold for our case:

\begin{itemize}[leftmargin=.7cm, labelsep=1em, itemsep=-1pt]
  \item[$\blacktriangleright$] \textcolor{black}{\textbf{Data is uniformly distributed on Riemannian manifold.} As the distribution of HOPE embeddings essentially depends on the connectivity pattern of the graph, for HP loss landscapes, we can expect this assumption to hold. This is because the number of neighbors for each node in the graph would largely stay the same based on our neighborhood definition. Therefore, there will be no significance variation in \textit{density} across the graph, and the HOPE embeddings in turn should approximately have a uniform distribution.}
  \item[$\blacktriangleright$] \textcolor{black}{\textbf{The Riemannian metric is locally constant.} The distances between HOPE embeddings correlate directly with the local connectivity patterns within the network. Given the stability in neighborhood structures as we discussed above, it is reasonable to presume that local distance metrics would remain approximately constant in small regions.}
  \item[$\blacktriangleright$] \textcolor{black}{\textbf{The manifold is locally connected.} HOPE embeddings can actually preserve more than just the local structure between configurations, as it is able to capture high-order proximities in the graph. We thereby expect this assumption will also hold.}
\end{itemize}

\textcolor{black}{\textit{\textbf{Remark on algorithmic choices.}} In principle, other node embedding and dimensionality reduction methods can be applied to serve our purpose. Our specific choice on HOPE is mainly because of its scalability to large-scale networks and its ability to preserve both local and global structure of the landscape. As for dimensionality reduction, \cite{DraganovJSMABA23} has made detailed theoretical comparisons between UMAP and $t$-SNE, and shows that only the normalization significantly impacts their outputs. This then implies that a majority of the algorithmic differences can be toggled without affecting the embeddings. We choose UMAP here for its better scalability. At the end, the quality of the visualization using our method will continuously grow with the new state-of-the-art in graph represention learning and dimensionality reduction.}

\textcolor{black}{\textit{\textbf{Remark on UMAP hyperparameters.}} Since HP loss landscapes can vary a lot in terms of dimensionality and total number of configurations, there is no `one-size-fits-all' setup for HPs of UMAP. Yet, here we provide some empirical guidelines for HP tuning, which mainly focus on two most important HPs, namely \texttt{n\_neighbors} and \texttt{min\_dist}. Specifically, \texttt{n\_neighbors} controls the balance between local and global structure in the embedding. In general, we found that it is better to set \texttt{n\_neighbors} to a value that is larger than the average number of neighbors for each node. For \texttt{min\_dist}, which specifies the minimum distance between points in the low-dimensional space, we recommend values larger than $0.5$. The reasoning here is that we want the points to be more spread out in the low-dimensional space, and thus prevent the points to be densely distributed in local regions. However, a \texttt{min\_dist} that is too large can also cause problems, as different parts of the landscape can get intertwined with each other.}

\subsection{Landscape Analysis Metrics}
\label{app:app_method_metrics}

\begin{algorithm}[t]
  \caption{Best-Improvement Local Search}
  \small
  \begin{algorithmic}[1] 
  \REQUIRE A starting configuration $\mathbf{c}$; A neighborhood function $\mathcal{N}_d$; A fitness function $f$
  \WHILE{$\mathbf{c}$ is not a local optimum}
      \STATE $\mathbf{c}^\prime_{\text{best}} = \arg\min_{\mathbf{c}^\prime \in \mathcal{N}_1(\mathbf{c})}f(\mathbf{c}^\prime)$
      \IF{$f(\mathbf{c}^\prime_{\text{best}}) < f(\mathbf{c})$}
          \STATE $\mathbf{c} \gets \mathbf{c}^\prime_{\text{best}}$
      \ENDIF
  \ENDWHILE
  \STATE \textbf{return} $\mathbf{c}$
  \end{algorithmic}
  \label{alg: localsearch}
\end{algorithm}

\begin{algorithm}[t]
  \caption{Identifying Local Optima and Their Basins}
  \small
  \begin{algorithmic}[1]
  \REQUIRE The set of all configurations $\mathcal{C}$
  \STATE $\mathcal{V} \gets \emptyset$ 
  \STATE $\mathcal{B} \gets \emptyset$ 
  \FORALL{$\mathbf{c} \in \mathcal{C}$}
      \STATE{$\mathbf{c}^\ell \gets \textsc{LocalSearch}(\mathbf{c})$}
      \STATE $\mathcal{B}[\mathbf{c}^\ell] \gets \mathcal{B}[\mathbf{c}^\ell] \cup \{\mathbf{c}\}$
  \ENDFOR
  \STATE \textbf{return} $\mathcal{V}, \mathcal{B}$
  \end{algorithmic}
  \label{alg: basin}
\end{algorithm}

\textbf{Assortativity coefficient.} The assortativity coefficient of a network assesses the degree to which nodes tend to be connected to other nodes that are similar w.r.t. some attributes. For example, in a social network, this would mean that people tend to be friends with other people who are similar to themselves in terms of education level, income, race\footnote{As one can image, the exact definition of assortativity can depend on whether the target attribute is categorical (i.e., unordered) or numerical (i.e., ordered). Here we focus on the latter as model loss is real-valued.}, etc. For HP loss landscape, the performance assortativity measures the extent to which HP configurations with similar performance are more likely to be neighbors to each other. Formallly, given a HP loss landscape as a directed graph, and the model loss $\mathcal{L}$ takes values $[L_1, L_2, \dots]$, the $\mathcal{L}$-assortativity evaluates the Pearson correlation of the measured loss between pairs of linked configurations and is measured as~\cite{Newman03}:
\begin{equation}
    \mathcal{L}_{\text{ast}}=\frac{\sum_{i}e_{ii}-\sum_{i}a_ib_i}{1-\sum_{i}a_ib_i}
\end{equation}
where $e_{ij}$ is called mixing matrix entry, which represents the fraction of total edges in the network (i.e., landscape) that connects configurations having performance $\mathcal{L}(\bm{\lambda}) = L_i$ to configurations having attribute $\mathcal{L}(\bm{\lambda}) = L_j$. In directed networks like our case, this can be asymmetric, i.e., $e_{ij} \neq e_{ji}$.  In addition, $a_i = \sum_{j}e_{ij}$ is the portion of edges $(\bm{\lambda}_u,\bm{\lambda}_v)$ such that $\mathcal{L}(\bm{\lambda}_u) = L_i$ and $b_i = \sum_{j}e_{ji}$ is the portion of edges $(\bm{\lambda}_v,\bm{\lambda}_u)$ such that $\mathcal{L}(\bm{\lambda}_v) = L_i$. A high $\mathcal{L}_{\text{ast}}$ would imply that configurations with similar performance have strong tendancy to be connected and form local clusters.

\textbf{Landscape neutrality.} It is often in genetics that a mutation in a single position of a DNA sequence will only lead to negligible change in the expression. Such phenomenon is known as landscape \textit{neutrality} at a macro level, and for each sequence, this can be quantitatively measured by the \textit{mutational robustness}~\cite{PayneW19}, which is the probability for such non-effective mutation to happen among all its possible mutants. Similar ideas are also applicable to HPO, where altering certain HPs may only result in subtle performance shifts. In particular, we define two neighboring configurations to be \textit{neutral} if their respective performances differ less than a small fraction $\epsilon$. We choose $\epsilon=0.1\%$ in this paper, since changes below this threshold would make almost no practical meaning. We then define the \textit{neutral ratio}, denoted as $\nu(\bm{\lambda})$, of a configuration $\bm{\lambda}$ as the portion of its neutral neighbors in its neighborhood. The avarage neutrality of the whole landscape is then defined as:
\begin{equation}
    \bar{\nu} = \mathbb{E}[\nu(\bm{\lambda})] = \frac{1}{|\bm{\Lambda}|} \sum_{\bm{\lambda} \in \bm{\Lambda}} \nu(\bm{\lambda})
\end{equation}
\textbf{Neutrality distance correlation.} While neutrality generally characterizes the expected probability for neutral moves to occur in the whole landscape, it can actually vary across regions. In particular, it is important to investigate whether it is more likely to encounter neutral moves when approaching the global optimum, as in practice, we often find a diminishing gain when tuning towards the best-possible configuration. We quantitatively assess this using the neutrality distance correlation (NDC), which measures the Pearson correlation coefficient between the neutrality of a configuration $\bm{\lambda}$ and its distance to the global optimum, $\mathrm{d}(\bm{\lambda}, \bm{\lambda}^*)$ (\pref{eq:ndc_1}).  Specifically, for each adaptive walk in the landscape using best-improvement local search (\pref{alg: localsearch}) that can eventually approach $\bm{\lambda}^*$, we measure the Pearson correlation coefficient between $\Delta \mathcal{L}$ for each pair of consecutive configurations $(\bm{\lambda}_i, \bm{\lambda}_{i-1})$ ($i\geq2$), and $\mathrm{d}(\bm{\lambda}_i, \bm{\lambda}^*)$. We calculate NDC as the average across all such walks. 
\begin{equation}
  \text{NDC} = \rho_p[\nu(\bm{\lambda}), \mathrm{d}(\bm{\lambda}, \bm{\lambda}^*)]
  \label{eq:ndc_1}
\end{equation}
\textbf{Number of local optima.} A configuration $\bm{\lambda}$ is said to be a \textit{local optimum} if its performance is superior to any other configuration in its neighborhood, i.e., $\forall\bm{\lambda}\in{\mathcal{N}(\bm{\lambda}^\ell)}$, we have $\mathcal{L}(\bm{\lambda}^\ell)<{\mathcal{L}(\bm{\lambda})}$. For a \textit{unimodal} landscape, there is only a global optimum configuration $\bm{\lambda}^*$. In constrast, multimodal landscapes have various local optima with sub-optimal performance, and is known to be `rugged'.

\textbf{Size of basin of attraction.} While a multimodal landscape can be difficult to optimize due to the pressence of various local optima, not all of them are equal in terms of the capability of trapping a solver. For a $2$D minimization scenario, this can be envisioned by the fact that each local optimum is located at the bottom of a `basin' in the landscape surface. Configurations in each basin would eventually fall into the corresponding basin bottom (i.e., local optimum) when applying a simplest hill-climbing local search (\pref{alg: localsearch}). The effort needed to escape from such a basin is direclty related to its sizes (e.g., depth and radius in a $2$D space). A local optimum with small basin size is very unlikely to cause significant obstacles to optimization, whereas the opposite is true for those featuring a dominated size of basin that is even larger than the global optimum. Formally, we define the \textit{basin of attraction} $\mathcal{B}$ of a local optimum $\bm{\lambda}^\ell$ to be the set of all configurations from which local search converges to $\bm{\lambda}^\ell$, i.e., $\mathcal{B}=\{\bm{\lambda}\in{\bm{\Lambda}}\mid\mathrm{LocalSearch}(\bm{\lambda})\rightarrow\bm{\lambda}^\ell\}$ (\pref{alg: basin}). The size of $\mathcal{B}$, denoted as $s_{\mathcal{B}}$, is defined as the cardinality of the basin set as $|\mathcal{B}|$. In this study, we report the mean basin size $\bar{s}_{\mathcal{B}}$ of all local optima (except the global optimum) in the landscape.

\textbf{Autocorrelation.} A common metric for characterizing the smoothness of a landscape is the autocorrelation $\rho_{a}$ on a series of performance values $\mathcal{L}$. These values are extracted for configurations in a random walk $\text{RW}=\{\bm{\lambda}_0, \bm{\lambda}_1, \dots, \bm{\lambda}_n\}$ in the search space $\bm{\Lambda}$. Formally:
\begin{equation}
    \rho_{a}(k) = \frac{\mathbb{E}[(\mathcal{L}(\bm{\lambda}_i)-\bar{\mathcal{L}})(\mathcal{L}(\bm{\lambda}_{i+k})-\bar{\mathcal{L}})]}{\mathbb{V}(\mathcal{L}(\bm{\lambda}_i))}, \forall \bm{\lambda}_i \in \bm{\Lambda}
\end{equation}
Here, $k$ represents a lag or a step difference in the indices of configurations, and in our case we consider $k=1$ since each step in our search grids have been specifically designed to mimic the tunning strategy commonly used by human experts. For each landscape, we conduct $100$ random walks of length $100$ and average $\rho_{a}$ across all measurements to mitigate the effects of randomness.

\subsection{Local Optima Network (LON)}
\label{app:lon}

\textcolor{black}{Beyond the number of local optima and their basin sizes, a further landscape property to interrogate is the \textit{distribution} of local optima and the \textit{connectivity} pattern between them. For example, an important question that we may concern is \textit{whether we can escape from a given local optimum to the global optimum?}, if yes, then, \textit{what is the chance of this?} Local optima networks (LON)~\cite{OchoaTVD08, VerelOT11}, which are rooted in the study of energy landscapes in chemical physics~\cite{Stillinger95}, address these questions by constructing a subspace of the original fitness landscape where the nodes indicate local optima, and edges represent possible transitions between them. In particular, an improving edge can be traced from local optimum configuration $\bm{\lambda}^\ell_i$ to $\bm{\lambda}^\ell_j$ if configurations in $\mathcal{B}(\bm{\lambda}^\ell_i)$ can escape to $\bm{\lambda}^\ell_j$ by applying a $2$-bit perturbation followed by local search. The edge weights $w_{i,j}$ indicate the total probability for such transitions to happen (see~\pref{alg: lon}) between two local optima. By conducting network mining on LONs (e.g., \cite{HuangL23}), we can get further insights into the distribution and connection between local optima as well as how they are potentially linked with the global optimum.}

\begin{algorithm}[t]
  \caption{Constructing Local Optima Network}
  \small
  \begin{algorithmic}[1] 
  \REQUIRE The set of local optima $\mathcal{V}$; The basin of attraction of each local optimum $\mathcal{B}_{\mathbf{c}^\ell}$; The set of all configurations $\mathcal{C}$; A neighborhood function $\mathcal{N}_d(\mathbf{c})$
  \STATE $\mathcal{E} \gets \emptyset$ 
  \STATE $\mathcal{W} \gets \emptyset$ 
  \FORALL{$\mathbf{c}^\ell \in \mathcal{V}$}
          \FORALL{$\mathbf{c}^{\ell\prime} \in \mathcal{N}_2(\mathbf{c}^\ell)$}
              \STATE{$\mathbf{c}^\ell_{\text{new}} \gets \textsc{LocalSearch}(\mathbf{c}^{\ell\prime})$}
              \IF{$f(\mathbf{c}^\ell_{\text{new}}) < f(\mathbf{c}^\ell)$}
                  \IF{edge $(\mathbf{c}^\ell, \mathbf{c}^\ell_{\text{new}})$ not in $\mathcal{E}$}
                      \STATE $\mathcal{E} \gets \mathcal{E} \cup \{(\mathbf{c}^\ell, \mathbf{c}^\ell_{\text{new}})\}$
                      \STATE $\mathcal{W}[(\mathbf{c}^\ell, \mathbf{c}^\ell_{\text{new}})] \gets 1$
                  \ELSE
                      \STATE $\mathcal{W}[(\mathbf{c}^\ell, \mathbf{c}^\ell_{\text{new}})] \gets \mathcal{W}[(\mathbf{c}^\ell, \mathbf{c}^\ell_{\text{new}})] + 1$
                  \ENDIF
              \ENDIF
      \ENDFOR
  \ENDFOR
  \STATE \textbf{return} $\mathcal{G} = (\mathcal{V}, \mathcal{E}, \mathcal{W})$
  \end{algorithmic}
  \label{alg: lon}
\end{algorithm}

\subsection{Landscape Similarity Metrics}
\label{app:app_method_sim}

\textbf{Spearman correlation.} It is a non-parametric measure of rank correlation which assesses how well the relationship between configuration performances in two landscapes can be described using a monotonic function. It is defined as the Pearson correlation coefficient between the performance ranks configurations in two landscapes.

\textbf{Shake-up metric.} It originates from the Kaggle competition community and is designed to assess the rank changes between the public and private leaderboards of a competition~\cite{Trotman19}. To be specific, it quantifies the average movement of rankings from public board to the private board. For HP loss landscapes, this metric can indicate the expected rank shifts for a configuration when evaluating it on two different scenarios (e.g., change the dataset).
\begin{equation}
    \text{Shake-up} = \frac{1}{|\bm{\Lambda}|} \sum_{\bm{\lambda} \in \bm{\Lambda}} \frac{|R(\mathcal{L}_1(\bm{\lambda})) - R(\mathcal{L}_2(\bm{\lambda}))|}{|\bm{\Lambda}|}
\end{equation}
\textbf{$\gamma$-set similarity.} It is proposed in~\cite{WatanabeAOH23} to assess the similarity of two tasks using the ratio of their most proninent configurations. More specifically, for two HP loss landscapes, their $\gamma$-set similarity is defined as the ratio of the intersection of the top-$\gamma$ regions to the union of them. In this paper, we consider $\gamma=10\%$.

\section{HP Loss Landscapes of SVM}
\label{app:svm}

\begin{figure}[t!]
  \centering
    \includegraphics[width=\linewidth]{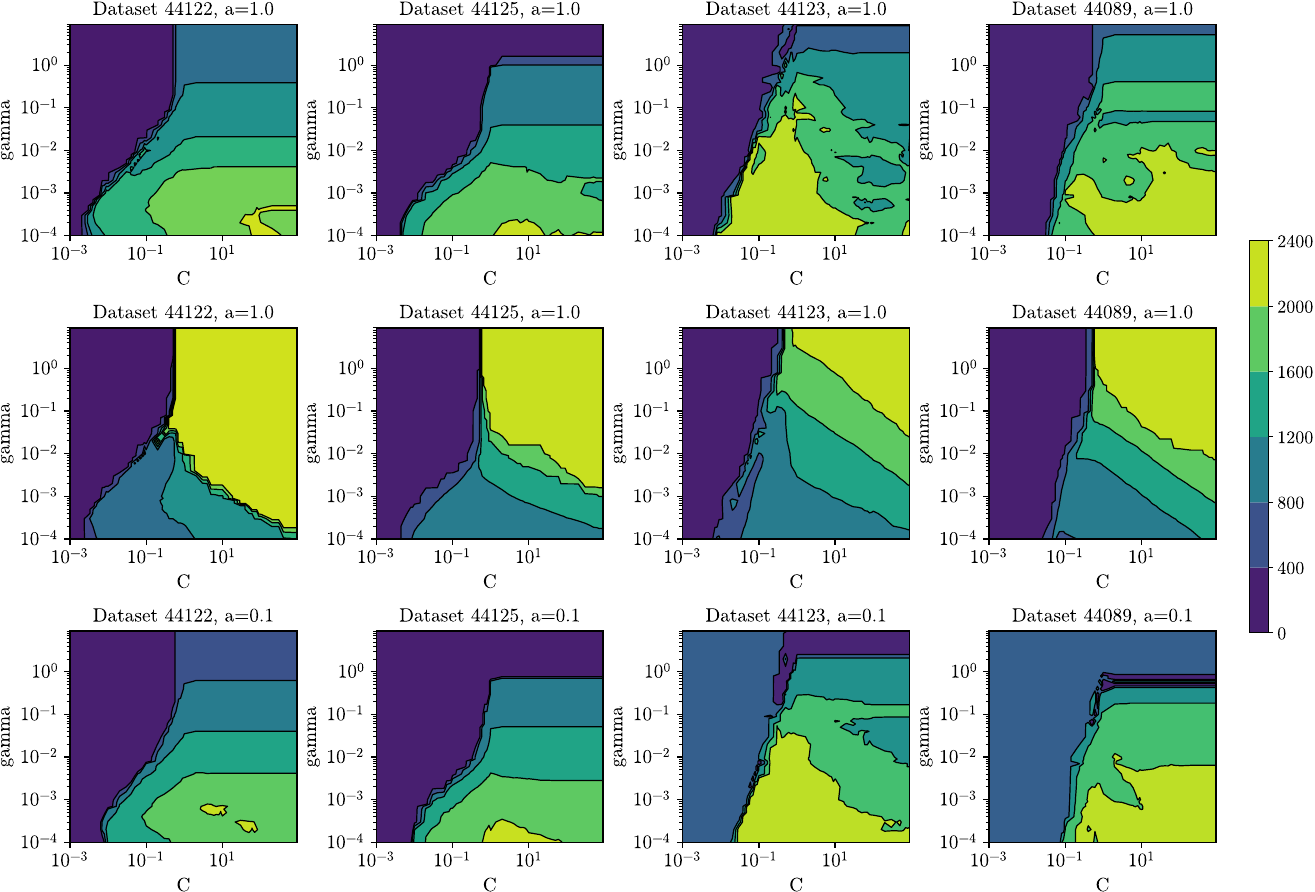}
    \caption{HP loss landscapes of \texttt{SVM} on four tabular datasets, with $\alpha=\{0.1, 1\}$ training data, with respect to training and test accuracy.}
    \label{fig:svm}
\end{figure}

Due to the inherent scalability issue of \texttt{SVM}, it is infeasible to evaluate it on most of our considered tabular datasets, as they often contains a large sample size and a high-dimensional feature space. Instead, we only conducted experiments on a few datasets that are relatively small in size with respective to two HPs (\texttt{gamma} and \texttt{c}) using RBF kernel. Since the resulting HP loss landscapes are already in a 2D space, we thus directly visualize them as heatmaps. \pref{fig:svm} presents the results on four tabular datasets (\#44122, \#44123, \#44089, \#44125) with $\alpha=\{0.1, 1\}$ training data. We can observe that the patterns here generally match our previous findings on other models, i.e., the landscapes are smooth, highly clustered, and nearly unimodal; a considerable portion of the landscape is neutral, and the performance difference diminishes around the optimum. These general characteristics also largely preserves across datasets, or with lower training fidelities. Nevertheless, similar to our observed patterns for \texttt{XGBoost}, here the training loss landscape exhibits very different topography compared to the test loss landscape, where overfitting is prevalent.

\section{Additional Results on Landscape Visualization.}

\subsection{Landscape Visualizations for DT, LGBM and RF}
\label{app:addi_visual}

\begin{figure}[t!]
  \centering
    \includegraphics[width=.75\linewidth]{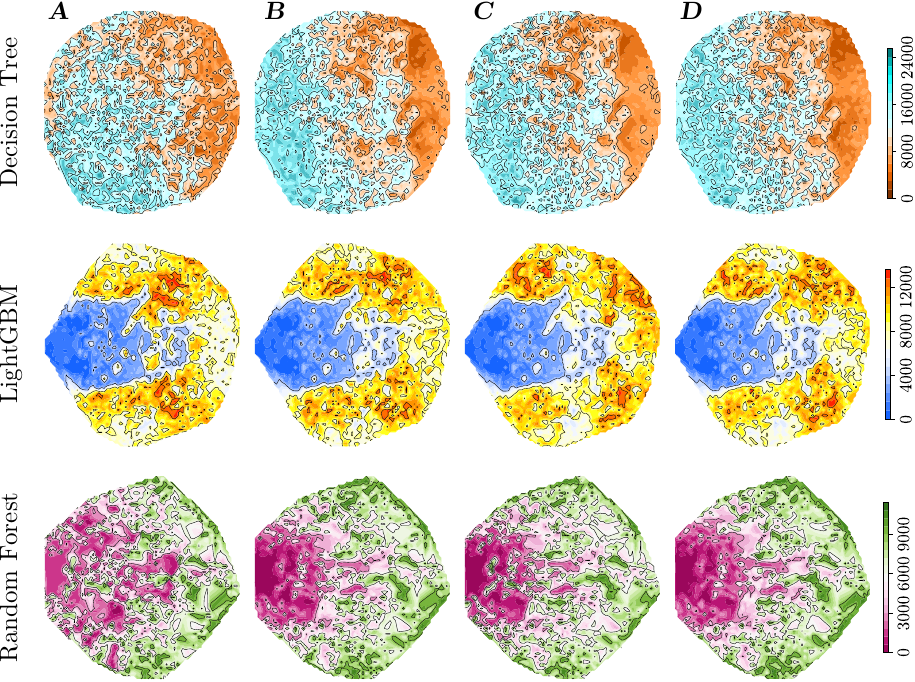}
    \caption{2D visualization of HP loss landscapes of \texttt{DT}, \texttt{LightGBM} and \texttt{RF} under different scenarios: \textbf{(A)} $\mathcal{L}_\text{test}$ landscape on baseline datasets (\#44059), \textbf{(B)} $\mathcal{L}_\text{train}$ landscape on baseline datasets, \textbf{(C)} Low-fidelity $\mathcal{L}_\text{test}$ landscape on baseline datasets obtained using $10\%$ training data, \textbf{(D)} $\mathcal{L}_\text{test}$ landscape on different datasets (\#44143). Colors indicate ranks of performance (lower rank values are better).}
    \label{fig:landscape_app}
\end{figure}

We present additional landscape visualizations for decision tree (\texttt{DT}), LightGBM (\texttt{LGBM}), and random forest (\texttt{RF}) in~\pref{fig:landscape_app}.

\subsection{Landscape Visualizations for NK-Landscape and BBOB Functions}
\label{app:nk}

\begin{figure}[t!]
  \centering
    \includegraphics[width=\linewidth]{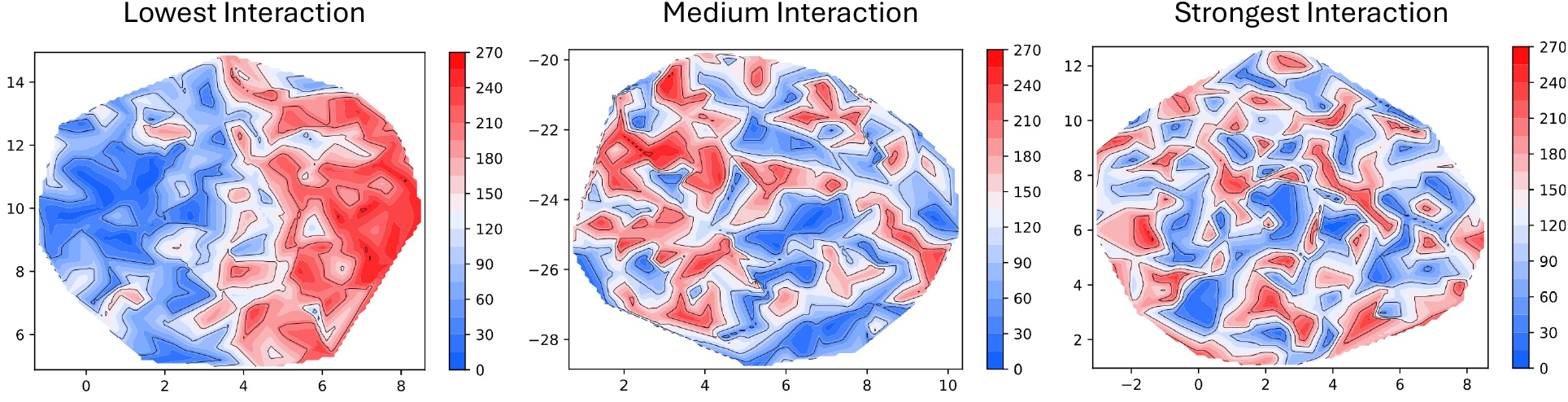}
    \caption{Visualizations of Kauffman's $NK$-landscape of $n=10$ with different levels of ruggedness using our proposed method in~\pref{sec:methodology} (low: $k=0$, medium: $k=5$, and strong: $k=10$).}
    \label{fig:landscape_nk}
\end{figure}

\begin{figure}[t!]
  \centering
    \includegraphics[width=\linewidth]{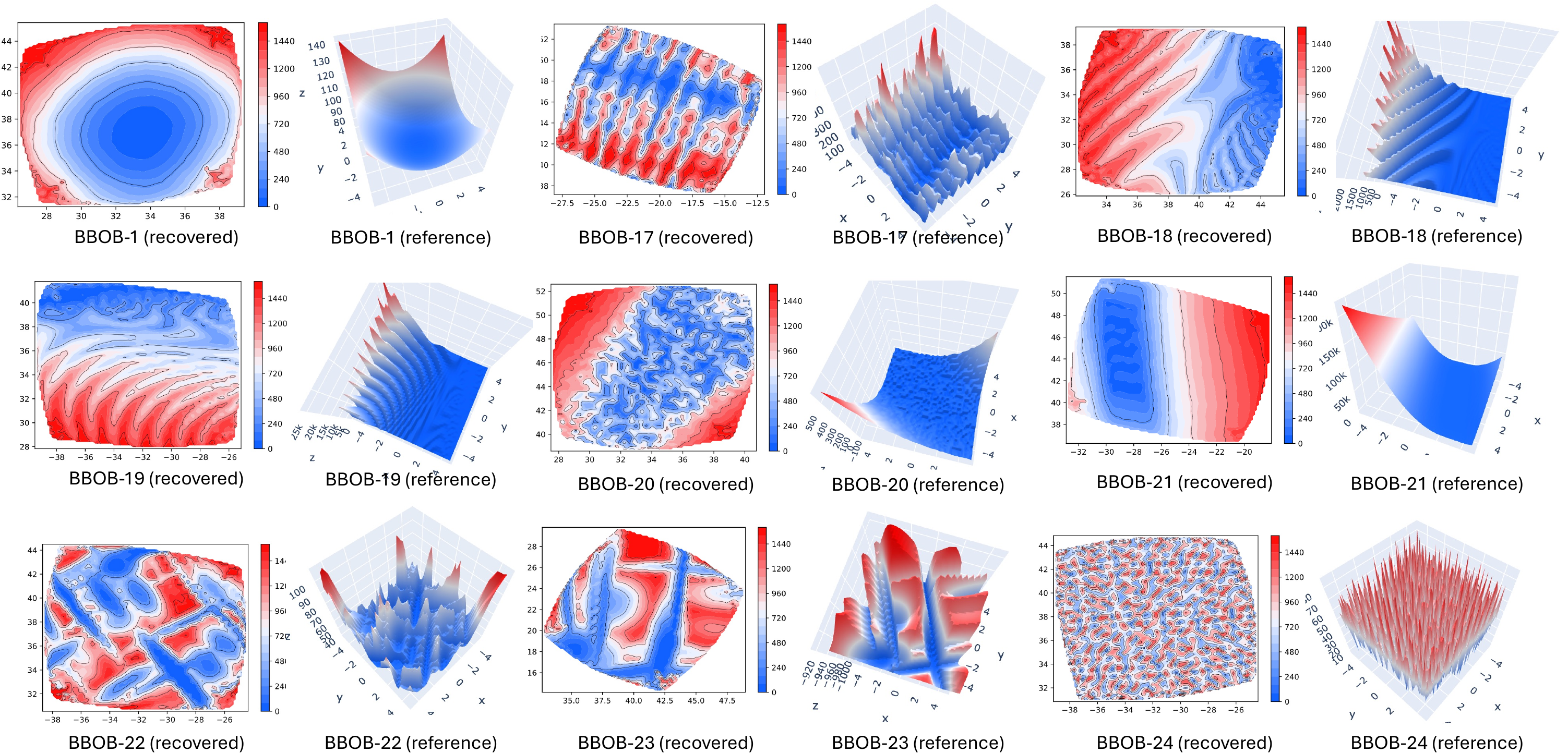}
    \caption{Recovered landscapes with our proposed visualization method versus groundtruth landscape (reference) surfaces of $f_1$ and $f_{17}$ to $f_{24}$ in the black-box optimization benchmark (BBOB) benchmarking suite from \texttt{COCO}.}
    \label{fig:landscape_bbob}
\end{figure}

In order to demonstrate the capability of our proposed landscape visualization method in capturing the structural characteristics of fitness landscapes, here we conduct additional experiments with artifical black-box optimization problems. 

\textbf{NK-landscape.} We first consider the well-known $NK$-landscape model~\cite{kauffman93}, which was introduced by Stuart Kauffman in the last century for studying the evolutionary of biological systems. It has two tunable parameters: $n$ and $k$. Here, $n$ represents the number of elements in the system, while $k$ determines the level of interaction between these elements. As $k$ increases, the landscape becomes increasingly rugged, with a greater number of local optima. For illustration here, we generate $NK$-landscapes with $n=10$ and three levels of ruggedness by varying the parameter $k\in \{0, 5, 10\}$. As shown in~\pref{fig:landscape_nk}, our visualization method is able to capture the increasing ruggedness of the landscapes as $k$ grows, where the landscape becomes more rugged.

\textbf{BBOB benchmark suite.} Another scenario we consider here is the black-box optimization benchmark (BBOB) functions suite from the \texttt{COCO} platform~\cite{HansenARMTB21}. This suite contains $24$ BBOB functions ($f_1$ to $f_{24}$) with a diverse range of complexities that are widely studied in literature. Here we test the capability of our visualization method in recovering the topography of these functions by comparing the visualized landscapes with the groundtruth surfaces. In particular, we present the results for $f_{17}$ to $f_{24}$ in \pref{fig:landscape_bbob}, which are known to exhibit strong multimodality (to compare, we also present $f_1$, a highly separable unimodal function). As can be seen from the results, our method is able to accurately recover shape of the function surface, even for the most complicated ones.

\section{Analysis on NAS Loss Landscapes}

\label{app:nas}

\subsection{NAS Benchmarks}
\label{app:nasbenchs}

\textcolor{black}{\textbf{NASBench-101.} It represents the first effort towards benchmarking NAS research and thus fostering reproducibility in the community. It evaluates a CNN architecture with $3$ stacked blocks, where a down-sampling is added between two consecutive blocks; A $3 \times 3$ convolution is used before the main blocks, and the outputs of the main blocks are fed to an average pooling and fully connected layer. NASBench-101 considers a \textit{cell}-based search space which enumerates all possible configurations for each block. More specifically, the search space is formulated as a directed acylic graph (DAG) with $7$ nodes and a maximum of $9$ edges. Here, each node can represent one of the following operations: $a)$: $1 \times 1$ convolution, $b)$: $3 \times 3$ convolution, and $c)$: max pooling. After removing all isomorphic cells, this search space results in $423$k unique configurations. NASBench-101 evaluates each of them on the CIFAR-10 dataset and records meta-data at the $\{4, 12, 36, 108\}^{\text{th}}$ epoch. }

\textcolor{black}{\textbf{NASBench-201.} It features a different skeleton compared to NASBench-101, in which a residual block is applied to connect $3$ cells. Each cell here is a DAG with $4$ nodes and $6$ edges. Morever, here, operations are represented by edges, which have $5$ types: $a)$: zeroize (do nothing), $b)$: $1 \times 1$ convolution, $c)$: $3 \times 3$ convolution, $d)$: $3 \times 3$ average pooling, $e)$: skip connection. The benchmark thus contains $5^6 = \textcolor{black}{15,625}$ unique model architectures, with each evaluated on $3$ different datasets: $i)$: \textcolor{black}{CIFAR-10}, $ii)$: \textcolor{black}{CIFAR-100}, $iii)$: \textcolor{black}{ImageNet-16-120} using \textcolor{black}{$200$} epochs. }

\textcolor{black}{\textbf{NASBench-360.}} Different from the previous two benchmarks, which aims at providing pre-evaluated performance data for NAS configurations, NASBench-360 rather offers a set of $10$ datasets from diverse domains for more comprehensive evaluations. Yet, the original paper also provides evaluations of the NASBench-201 search space on two tasks: $\blacktriangleright$ NinaPro: it considers classifying sEMG signals into 18 classes of hand gestures; $\blacktriangleright$ Darcy Flow: it considers predicting the final state of a fluid from its initial conditions. 

\subsection{NAS Landscape Construction}

\textcolor{black}{Despite the cell-based search spaces of NAS benchmarks are very different from the HP ones considered in this paper, our landscape construction rountine could be easily transfered to NAS by redefining the neighborhood structure. This then demands proper encoding of the NAS configurations and the definition of a suitable distance function.}

\textcolor{black}{\textbf{NASBench-101 Neighborhood.} In the original paper, a cell-encoding method based on adjacency matrices is introduced to encode configurations, which comprises two components. First, a $7 \times 7$ upper-triangular binary matrix is used to indicate whether an edge exist between two nodes and thus determine the connectivity pattern of nodes. Next, the functionality of a configuration also depends on which operation is performed at each node, and this could be encoded via a vector of length $5$ (the input and output nodes are the same across archiectures and are omitted, non-existent nodes are represented by \texttt{NaN}). Therefore, a configuration could be specified using an adjacency matrice and a node vector. We define a neighbor of a given configuration to be the one with $1$-edit distance from it (e.g., either adding or deleting one edge, or change the operation of one node), as in the original paper. Note that not all such configurations are valid in the benchmark, as they can be isomorphic to others. The benchmark API provides built-in function to check for this.}

\textcolor{black}{\textbf{NASBench-201/360 Neighborhood.} The encoding of configurations in this search space is much more straight forward. Specifically, we encode each configuration using a $6$-bit vector, where each bit specifies the operation taken at each edge, which can take $5$ categorical values. We base our neighborhood definition here on the $1$-edit distance as well, in which a $1$-bit mutant of a configuration is the one with the operation in only one edge altered.}

\subsection{Results on NAS Loss Landscapes}

\textcolor{black}{While we have conducted all analyses on both benchmarks, it would be too tedious to lay all the information here. Instead, we use NASBench-101 to discuss general NAS landscape characteristics and multi-fidelity, and present the comparisons between NAS landscapes across datasets using NASBench-201. Before we present our results, we also note that while the majority of configurations ($359$k out of $423$k) in NASBench-101 search space come with $7$ nodes (i.e., with $5$ intermediate operations), the rest of them have less nodes. Here we mainly focus on configuragtions with $7$ nodes, since accounting all the configurations would result in many independent components in the landscape. We do not expect this to significantly affect our results.}

\textcolor{black}{\textbf{NAS Landscape Visualization.} We first visulize the NASBench-101 landscape using our proposed landscape visualization method in~\pref{sec:methodology}, as shown in~\pref{fig:landscapes_all} (C). Specifically, we plot the training accuracy landscape along with test accuracy landscapes trained at $4$ different number of epochs. It is clear to see from the plot that the landscape is far from unimodal, with many local optima. However, configurations still tend to form local clusters, though the relative size of each plateau seems to be much smaller than we see for HP loss landscapes. Considering that the search space here contains nearly $30$ times more confiurations than the HP space we used in the main text, each cluster may contain thousands of configurations, inside which the landscape could still be sufficiently smooth. }

% \begin{figure*}[t!]
%   \centering
%     \includegraphics[width=\linewidth]{figs/nasbench101.pdf}
%     \caption{\textcolor{black}{Visualization of NASBench-101 landscape using our proposed method. Here, we present the training accuracy landscapes as well as test accuracy landscapes at different recorded epochs. Color indicates rank of test accuracy, and higher values are better.}}
%     \label{fig:nasbench101}
% \end{figure*}

\textcolor{black}{\textbf{NAS Landscape Metrics.} Here we report several landscape metrics for the NASBench-101 landscape with respect to test accuracy at the $108$\textit{-th} epoch:}
\begin{itemize}[leftmargin=.7cm, labelsep=1em, itemsep=-0.5pt]
  \item \textcolor{black}{\textbf{Autocorrelation.} We obtained a correlation coefficient of $0.6031$ on the landscape. This confirms our hypothesis above that the landscape is still sufficiently smooth and highly navigable, despite we observed more complex patterns in the visualizations.}
  \item \textcolor{black}{\textbf{Clustering.} The \textit{accuracy}-assrotativity is $0.6485$, which indicates a good level of local clustering of configurations with similar performance levels. This further confirms that the landscape is locally smooth.}
  \item \textcolor{black}{\textbf{Neutrality.} Our neutrality measure, however, only yields a value of $0.075$, and suggests that most $1$-bit changes in a configuration would result in performance shift $>0.1\%$.}
  \item \textcolor{black}{\textbf{NDC.} Despite the overall neutrality of the NAS landscape is low, we still observe a high NDC value of $0.7194$. This implies a strong plateau trend near the optimum, where optimizers would pay considerable more effort to gain marginal performance improvement.}
  \item \textcolor{black}{\textbf{Number of Local Optima.} One of the most distinguishable property of NASBench-101 is a large collection of local optima in the landscape. In fact, we found $5,908$ local optima in total (out of the $359$k configurations), which could probably make the landscape far more difficult to optimization. We will discuss more about them and their basins in the LON part. }
\end{itemize}

\begin{figure}[t!]
  \centering
    \includegraphics[width=.75\linewidth]{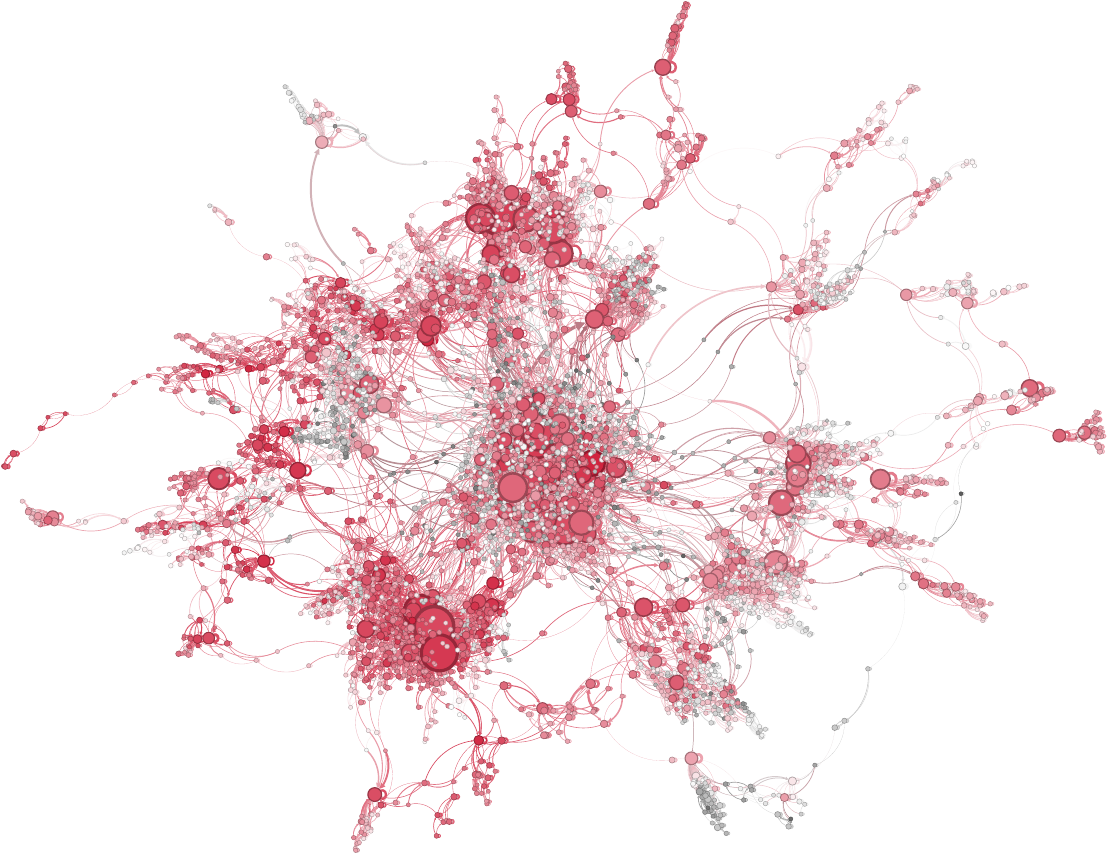}
    \caption{Visualization of the local optima network of NASBench-101 test landscapes at the $108$\textit{-th} epoch, containing $5,908$ nodes. Radius of each node (i.e., local optimum) indicates size of the corresponding basin of attraction. The color indicates test accuracy, where warmer color is better. Edges indicate transition probabilities between local optima, where thicker, warmer edges imply the corresponding transition if more likely to happen. Edge directions indicate the improving direction (i.e., pointing to the fitter configuration).}
    \label{fig:naslon}
\end{figure}

\textcolor{black}{\textbf{NAS Landscape Across Fidelities.} From~\pref{fig:landscapes_all} (C), we could see that in general the test landscapes with lower fidelities resemble the one trained with $108$ epoch. Quantitatively, the Spearman correlation between the test accuracy landscapes at $108$\textit{-th} epoch and $32$\textit{-th} epoch is $0.904$, with a Shake-up metric of $9.74\%$. These suggest a good general correlation between these landscapes, although there are $4$ times difference in their budget. When zooming into the $10\%$ region, the $\gamma$-set similarity is $0.64$, which is also farily good (the intesection ratio between top-$10\%$ regions is $78\%$). When we further decrease the budge by $4$ times, the Spearman correlation and Shake-up metric obtained between the $12$\textit{-th} and $108$\textit{-th} epoch are $0.657$ and $18.4\%$ respectively, while the $\gamma$-set similarity is $0.317$. Finally, with only $4$ epochs of training, the above metrics further change to $0.504$, $22.4\%$ and $0.164$ respectively. In general, the landscapes are still moderately correlated, but the detailed pattern can be largely distorted.}

% \begin{figure}[t!]
%   \centering
%     \includegraphics[width=.65\linewidth]{figs/nasbench201.pdf}
%     \caption{Visualization of NASBench-201 landscape using our proposed method across datasets.}
%     \label{fig:nasbench201}
% \end{figure}

\begin{figure}[t!]
  \centering
    \includegraphics[width=\linewidth]{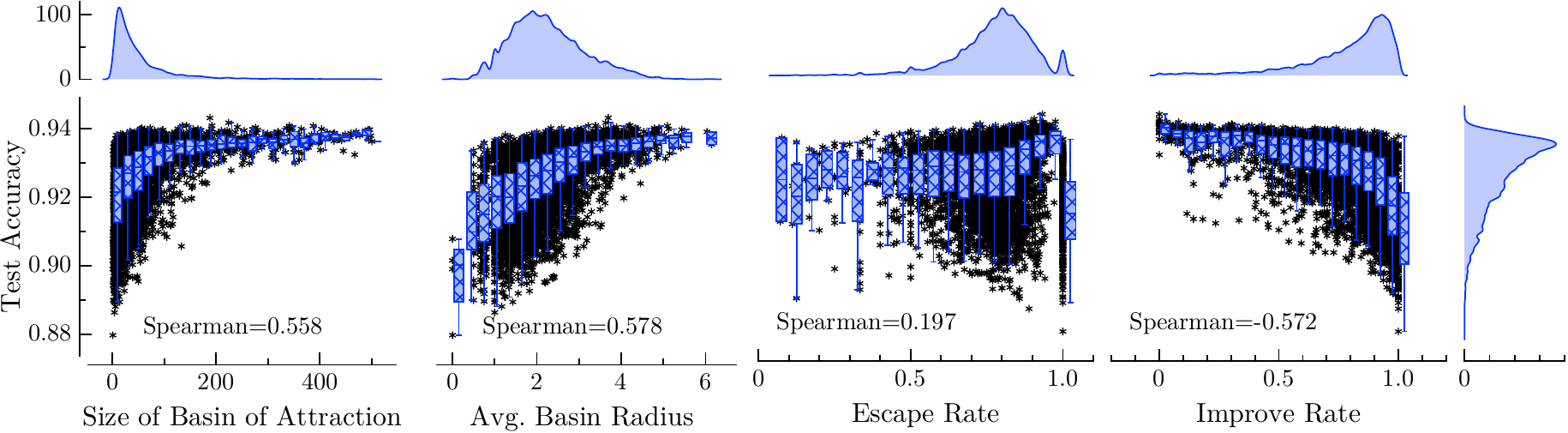}
    \caption{Distribution of local optima test accuracy versus LON metrics.}
    \label{fig:box}
\end{figure}

\begin{figure}[t!]
  \centering
  \includegraphics[width=.35\linewidth]{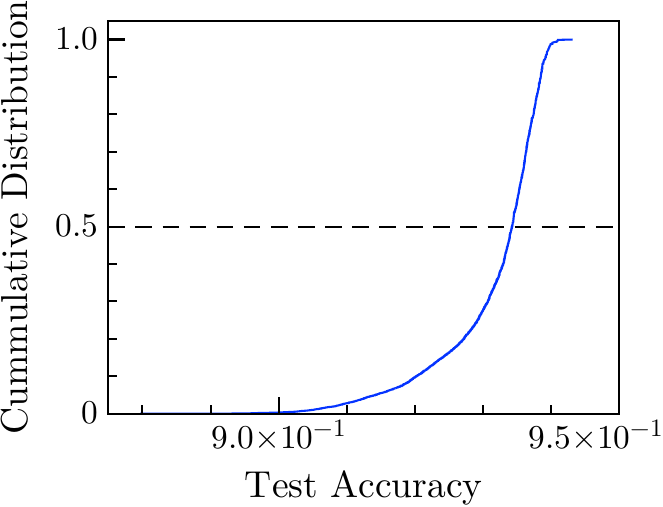}
  \caption{Cumulative distribution of basin size versus local optima performance.}
  \label{fig:cds}
\end{figure}

\textcolor{black}{\textbf{Local Optima Network Analysis.}} \textcolor{black}{\pref{fig:naslon} shows the local optima network of NASBench-101 test landscapes at the $108$\textit{-th} epoch. It could be obviously seen that there are clear community (clustering) structure in the network, where local optima with large basin of attractions tend to locate at the center of each cluster, which usually feature a promising performance. To be specific, from the left plots in~\pref{fig:box}, we could clearly observe that the size and radius\footnote{Here, we define the \textit{radius} of a basin of attraction as the expected number of local search steps needed to reach the corresponding local optimum. Since in practice, we observe that this value is often highly correlated with the size of basin of attraction, and correlate with performance in a similar manner, we only report basin size in the main text.} of the basin of attraction is positively correlated with the performance of local optima (Spearman correlation $>0.55$). More importantly, as suggested by the cumulative distribution of basin size shown in~\pref{fig:cds}, the highly-fit local optima have a dominant size of basin of attractions in the landscape. For example, the dashed line in~\pref{fig:cds} indicates that the cumulative sum of basin sizes of those local optima with $acc > 94.3\%$ take $50\%$ of the total basin size (which is the number of total configurations in the landscape). Since being in the basin of a local optimum would imply that local search will eventually converge it, our result is to say that if we start local search from a random configuration in the landscape, there is $50\%$ chance that we would end up in a local optimum with $acc > 94.3\%$. This is a very promising result, since we find such a level of accuracy is already better than $98.58\%$ of total configurations in the whole search space! It is also superior than $76.84\%$ of other local optima in the landscape.}

\textcolor{black}{A natural question that follows is the efforts that we need to reach such local optima. Statistically, we find that on average, it would take $3.04$ local search steps to reach a local optimum, while the mean of the longest walk length in each basin is $6.46$ steps. This is not a huge effort, since after taking such steps, as we discussed above, we have a good chance to fall into local optimum with $acc > 94.3\%$. However, we note that this is also not that trivial, since each `step' here means we exhaustively search for all the neighbors of a configuration, and then select the best one to move on. Such local search technique is called a \textit{best-improvement} local search. In contrast, there is also \textit{first-improvement} local search, in which we take the first configuration in the neighborhood that is fitter than the current one without considering other neighbors. This in general woul require more steps to reach a local optimum, but with fewer model evaluations at each step. While local search is definitely sub-optimal compared to advanced global search strategies, finding that even such a simple technique could somehow lead to a guaranteed promising result would imply that the NAS landscape is also benign.}

\textcolor{black}{Beyond local search, we then continue to consider if we do fall in a local optima whose fitness is not that satisfactory, then \textit{what is the chance that we can somehow escape from it?}. The two plots at the right panel of~\pref{fig:box} show the distribution of \textit{escape rate} and \textit{improve rate} correlate with test accuracy. Here, the escape rate is the chance that, a configuration in the basin of a local optimum $\mathbf{c}^\ell$, after applying a $2$-bit perturbation, would converge to a different local optimum $\mathbf{c}^{\ell}_{\text{new}}$. On the other hand, the improve rate further restricts that the new local optimum should have better performance compared to the current one. From~\pref{fig:box}, we could clearly see observe the majority of the local optima feature a escape rate of larger than $50\%$, which suggests that most local optima are not that difficult to escape from. For improve rate, we observe a good correlation with test accuracy, where it is more easy to find an improving move for a poorly-performed local optimum. Unfortunately, for those that already have promising performance, there is only little chance to transit to a better basin using $2$-bit perturbation.}

% \subsection{Results on NASBench-201}

% We visualize the loss landscapes of NASBench-201 on the $3$ datasets, namely CIFAR-10/100 and ImageNet, in~\pref{fig:nasbench201}. In general, we see that results on these $3$ tasks reveal strong consistency to each other (Spearman correlation $>0.95$), which conforms with our findings on HP loss landscapes.

\section{Landscape Similarity \& Datasets Characteristics}
\label{app:dataset_size}

\begin{figure}[t!]
  \centering
    \includegraphics[width=.75\linewidth]{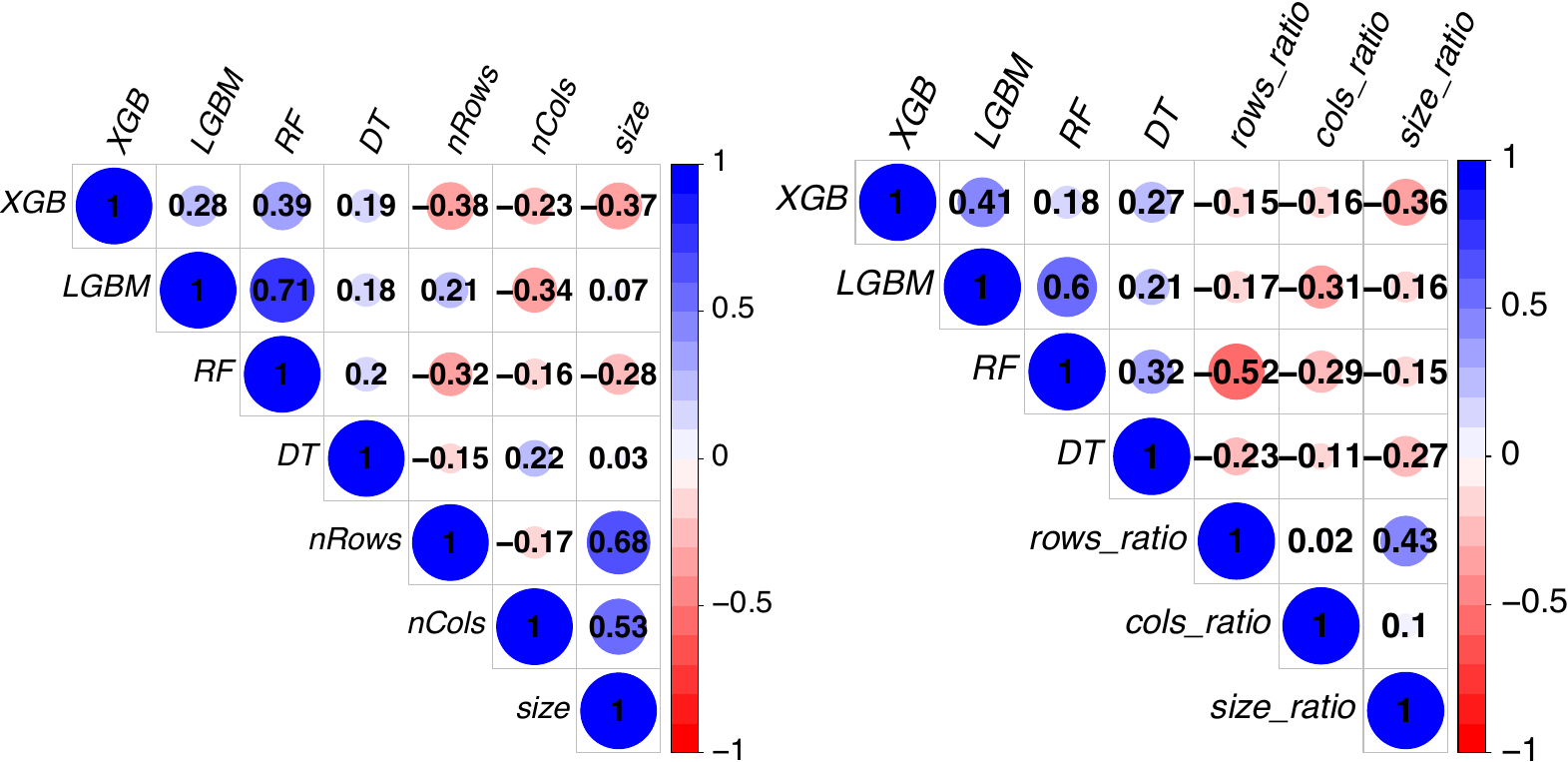}
    \caption{Spearman correlation between landscape similarity and dataset characteristics. Left: similarity between training and test loss landscapes for different models versus dataset characteristics. Right: similarity between landscape on different dataset for different models versus ratio of dataset characteristics. Here, dataset \textit{size} is defined as the product of the number of instances and features.}
    \label{fig:dataset_sim}
\end{figure}

\textcolor{black}{While in general, we find that HP loss landscapes studied in this paper share various common characteristics, the detailed topography can vary with dataset that the model is trained on. How characteristics of datasets could affect the landscape is an interesting problem to explore in more detail. In addition, we also hypothesize that the generalization gap of a model can also depend on the dataset, in addition to its HP setting. To investigate these, we conduct additional experiments on the $57$ tabular datasets to analyze the relationship between landscape similarity and dataset characteristics.}

\textcolor{black}{We first find in the left plot of~\pref{fig:dataset_sim} that the similarity between test and training loss landscapes is positively correlated across models. It implies that on certain datasets, all the $4$ models are more prone to overfit, whereas the opposite could be true for other datasets. This then verifies our hypothesis that overfitting not only depends on the model HPs, but also the dataset itself. To further explore which properties of the dataset could potentially contribute to this, we investigated the correlation between train-test landscape similarity and the number of instances \& features of each dataset, and their product. The results (also in~\pref{fig:dataset_sim} (left)) indicate that the degree of overfitting is correlated with all $3$ dataset size measures. In particular, for most models, it would be more likely to encounter overfitting on larger datasets. }

\textcolor{black}{We then proceed to investigate whether correlations between HP loss landscapes induced on different datasets are related to the relative size of the datasets. From~\pref{fig:dataset_sim} (right), it is clear to see that pairwise landscape similarities are obviously correlated across models, implying that all the models are likely to induce very different (or the opposite) landscapes on certain pairs of datasets. We could also see that these pairwise similarities are again correlated with the differences in dataset sizes. In particular, for datasets that have very different sizes, the resulting HP loss landscapes also tend to be different from each other.}

\textcolor{black}{However, we note that the correlations reported in both scenarios are somewhat weak, and the reasons could be two folds. First, the choice of datasets is only a partial factor contributing to landscape similarity, many other factors like model HPs, training settings can also important roles here. Second, the dataset meta-features we used here are rather naive, and there could be more comprehensive features for characterizing dataset properties, e.g.,~\cite{FeurerSH15} had used a collection of $46$ features to assess dataset similarity. However, this is beyond the scope of this work and we leave it to future works.}

\section{Details of the Experimental Setup}
\label{app: appendix_setup}

\subsection{Search Spaces}
\label{app:appendix_search_space}

\textcolor{black}{In this subsection, we elaborate the principles that we follow in designing the search spaces for each model. We also provide the detailed hyperparameter grid space for them in~\pref{tab:space_xgb} (\texttt{XGBoost}),~\pref{tab:space_dt} (\texttt{DT}),~\pref{tab:space_rf} (\texttt{RF}),~\pref{tab:space_lgbm} (\texttt{RF}),~\pref{tab:space_svm} (\texttt{SVM}), and~\pref{tab:space_cnn} (\texttt{CNN}). The top-level principle we follow is to include commonly used HPs and exclude unimportant ones, while keeping a good balance between search space coverage and computational cost. We first determine the list of HPs to be considered, and to this end:}

\begin{itemize}[leftmargin=.7cm, labelsep=1em, itemsep=-1pt]
  \item \textcolor{black}{We surveyed commonly used HPs in the practice (e.g., \cite{RijnH17} identified important HPs of several models using large-scale meta-data from OpenML) and HPO literature (e.g., the search spaces used in HPOBench~\cite{EggenspergerMMF21}). For \texttt{CNN}, we also refer to the design of NAS search spaces~\cite{ChittyVenkataEVS23}.}
  \item \textcolor{black}{We also run preliminary experiments by fixing all except but one HP to their default values, and vary a single HP using a wide range of values. This allows us to estimate the influence of each HP on model performance without conducting large-scale search. }
\end{itemize}

\textcolor{black}{We then combine the knowledge obtained in these precedures to design a rough HP list to be studied for each model. We then proceed to determine the domain and granularity of each HP, for which we bear the following considerations:}

\begin{itemize}[leftmargin=.7cm, labelsep=1em, itemsep=-1pt]
  \item \textcolor{black}{The domain of each HP should be large enough to cover the range of values that are commonly used in practice, while keep a balance with the computational cost. For example, we can not afford to search for \texttt{XGBoost} with thousands of base learners.}
  \item \textcolor{black}{Given the limited computational budget, there should be more number of bins for HPs that are more important, e.g., learning rates.}
  \item \textcolor{black}{Given the limited computational budget, we would prefer slightly removing $1$ or $2$ less important HPs, than having a search grid with many real-valued HPs have only $2$ to $3$ bins. }
\end{itemize}

\textcolor{black}{Following these principles, our final search spaces generally contain $5$ to $8$ HPs, with total number of configurations ranging from $6,480$ to $24,200$. We consider these search spaces representative of the real-world practice and thus form a good basis for landscape analysis.}

\begin{table*}[htbp]
  \centering
  \caption{\texttt{XGBoost} HP Grid Space ($14,960$ Configurations)}
    \small
    \begin{tabular}{||llr||}
    \hline
    \rowcolor{darkgray!30}Hyperparameter & Grid Values & Count \\
    \hline
    \hline
    \texttt{learning\_rate} & $[1\mathrm{e}^{-3}, 1\mathrm{e}^{-2}, 3\mathrm{e}^{-2}, 5\mathrm{e}^{-2}, 7\mathrm{e}^{-2}, 1\mathrm{e}^{-1}, \dots 5\mathrm{e}^{-1}]$ & $11$\\
    \rowcolor{darkgray!15}\texttt{subsample} & $[0.2, 0.4, 0.6, 0.8, 1]$ & $5$\\ 
    \texttt{max\_depth} & $[4, 5, 6, \dots, 20]$  & $17$\\ 
    \rowcolor{darkgray!15}\texttt{max\_bin} & $[256, 512, 1024, 2048]$  & $4$\\
    \texttt{n\_estimators} & $[100, 200, 300, 500]$ & $4$\\ 
    \hline
    \end{tabular}
  \label{tab:space_xgb}
\end{table*}

\begin{table*}[htbp]
  \centering
  \caption{\texttt{DT} HP Grid Space ($24,200$ Configurations)}
    \small
    \begin{tabular}{||llr||}
    \hline
    \rowcolor{darkgray!30}Hyperparameter & Grid Values & Count \\
    \hline
    \hline
    \texttt{splitter} & [\texttt{"best"}, \texttt{"random"}] & $2$\\ 
    \rowcolor{darkgray!15}\texttt{min\_samples\_split} & $[2, 4, \dots, 20]$ & $10$\\ 
    \texttt{max\_samples\_leaf} & $[1, 2, 4, \dots, 20]$  & $11$\\ 
    \rowcolor{darkgray!15}\texttt{max\_depth} & $[5, 10, \dots, 50, \texttt{None}]$  & $11$\\ 
    \texttt{max\_features} & $[0.1, 0.2, \dots, 1.0]$  & $10$\\ 
    \hline
    \end{tabular}
  \label{tab:space_dt}
\end{table*}

\begin{table*}[htbp]
  \centering
  \caption{\texttt{RF} HP Grid Space ($11,250$ Configurations)}
    \small
    \begin{tabular}{||llr||}
    \hline
    \rowcolor{darkgray!30}Hyperparameter & Grid Values & Count \\
    \hline
    \hline
    \texttt{min\_samples\_split} & $[2, 5, 10, 15, 20]$ & $5$\\ 
    \rowcolor{darkgray!15}\texttt{max\_samples\_leaf} & $[1, 5, 10, 15, 20]$  & $5$\\ 
    \texttt{Bootstrap} & [\texttt{True}, \texttt{False}] & $2$\\ 
    \rowcolor{darkgray!15}\texttt{max\_features} & $[0.2, 0.4, \dots, 1.0]$  & $5$\\ 
    \texttt{n\_estimators} & $[50, 100, 150, 200, 300]$ & $5$\\       
    \rowcolor{darkgray!15}\texttt{max\_depth} & $[10, 15, \dots, 50, \texttt{None}]$  & $9$\\ 
    \hline
    \end{tabular}
  \label{tab:space_rf}
\end{table*}

\begin{table*}[htbp]
  \centering
  \caption{\texttt{LightGBM} HP Grid Space ($13,440$ Configurations)}
    \small
    \begin{tabular}{||llr||}
    \hline
    \rowcolor{darkgray!30}Hyperparameter & Grid Values & Count \\
    \hline
    \hline
    \texttt{learning\_rate} & $[1\mathrm{e}^{-2}, 5\mathrm{e}^{-2}, 7\mathrm{e}^{-2}, 1\mathrm{e}^{-1}, \dots 5\mathrm{e}^{-1},]$ & $10$\\ 
    \rowcolor{darkgray!15}\texttt{boosting\_type} & [\texttt{"gbdt"}, \texttt{"dart"}] & $2$\\ 
    \texttt{max\_depth} & $[5, \dots, 30, \texttt{None}]$  & $16$\\ 
    \rowcolor{darkgray!15}\texttt{n\_estimators} & $[50, 75, 100, 150, 200, 250, 300]$ & $7$\\     
    \texttt{num\_leaves} & $[10, 20, 30, 40, 50, 100]$ & $6$\\ 
    \hline
    \end{tabular}
  \label{tab:space_lgbm}
\end{table*}

\begin{table*}[htbp]
  \centering
  \caption{\texttt{SVM} HP Grid Space ($2340$ Configurations)}
    \small
    \begin{tabular}{||llr||}
    \hline
    \rowcolor{darkgray!30}Hyperparameter & Grid Values & Count \\
    \hline
    \hline
    \texttt{gamma} & $[1\mathrm{e}^{-4}, 10]$ & $45$\\ 
    \rowcolor{darkgray!15}\texttt{c} & $[1\mathrm{e}^{-3}, \dots, 1\mathrm{e}^{3}]$ & $54$\\ 
    \hline
    \end{tabular}
  \label{tab:space_svm}
\end{table*}

\begin{table*}[htbp]
  \centering
  \caption{\texttt{CNN} HP Grid Space ($6,480$ Configurations)}
    \small
    \begin{tabular}{||llr||}
    \hline
    \rowcolor{darkgray!30}Hyperparameter & Grid Values & Count \\
    \hline
    \hline
    \texttt{act\_fn} & [\texttt{"gelu"}, \texttt{"relu"}, \texttt{"tanh"}] & $3$\\ 
    \rowcolor{darkgray!15}\texttt{batch\_norm} & [\texttt{True}, \texttt{False}] & $2$\\ 
    \texttt{learning\_rate} & [$1\mathrm{e}^{-4}, 5\mathrm{e}^{-4}, 1\mathrm{e}^{-3}, 3\mathrm{e}^{-3}, 5\mathrm{e}^{-3}$]  & $5$\\ 
    \rowcolor{darkgray!15}\texttt{batch\_size} & [$128, 256$]  & $10$\\ 
    \texttt{drop\_out} & [$0.0, 0.1, 0.25, 0.5$] & $4$\\ 
    \hline
    \rowcolor{darkgray!15}\texttt{width\_linear} & [$256, 512, 1024$]  & $3$\\ 
    \texttt{width\_conv} & [$32, 64, 128$] & $3$\\ 
    \rowcolor{darkgray!15}\texttt{\#conv\_block} & [$2, 4, 6$] & $3$\\ 
    \hline
    \end{tabular}
  \label{tab:space_cnn}
\end{table*}

\subsection{Datasets}
\label{app: appendix_datasets}

Here we provide basic information regarding the $5$ groups of datasets used for this study, including: $i)$: numerical regression (\pref{tab:num_reg}), $ii)$: numerical classification (\pref{tab:num_class}), $iii)$: categorical regression (\pref{tab:cat_reg}), $iv)$: categorical classification (\pref{tab:num_class}), and $v)$: image classifcation (\pref{tab:img_class}).

\begin{table*}[htbp]
  \centering
  \caption{Numerical regression}
    \small
    \begin{tabular}{||clrr||}
    \hline
    \rowcolor{darkgray!30}OpenML ID & Dataset Name & \#Samples & \#Features \\
    \hline
    \hline
    44132 & cpu\_act & $8,192$ & $21$ \\
    \rowcolor{darkgray!15}44133 & pol   & $15,000$ & $26$ \\
    44134 & elevators & $16,599$ & $16$ \\
    \rowcolor{darkgray!15}44136 & wine\_quality & $6,497$ & $11$ \\
    44137 & Ailerons & $13,750$ & $33$ \\
    \rowcolor{darkgray!15}45032 & yprop\_4\_1 & $8,885$ & $42$ \\
    44138 & houses & $20,640$ & $8$ \\
    \rowcolor{darkgray!15}44139 & house\_16H & $22,784$ & $16$ \\
    45034 & delays\_zurich\_transport & $5,465,575$ & $9$ \\
    \rowcolor{darkgray!15}44140 & diamonds & $5,3940$ & $6$ \\
    44141 & Brazilian\_houses & $10,692$ & $8$ \\
    \rowcolor{darkgray!15}44142 & Bike\_Sharing\_Demand & $17,379$ & $6$ \\
    44143 & nyc-taxi-green-dec-2016 & $581,835$ & $9$ \\
    \rowcolor{darkgray!15}44144 & house\_sales  & $21,613$ & $15$ \\
    44145 & sulfur & $10,081$ & $6$ \\
    \rowcolor{darkgray!15}44146 & medical\_charges & $163,065$ & $5$ \\
    44147 & MiamiHousing2016 & $13,932$ & $14$ \\
    \rowcolor{darkgray!15}44148 & superconduct  & $21,263$ & $79$ \\
    \hline
    \end{tabular}
  \label{tab:num_reg}
\end{table*}

\begin{table*}[htbp]
    \centering
    \caption{Numerical classification}
      \small
      \begin{tabular}{||clrr||}
      \hline
      \rowcolor{darkgray!30}OpenML ID & Dataset Name & \#Samples & \#Features \\
      \hline
      \hline
      44120 & electricity & $38,474$ & $7$ \\
      \rowcolor{darkgray!15}44121 & covertype & $566,602$ & $10$ \\
      44122 & pol   & $10,082$ & $26$ \\
      \rowcolor{darkgray!15}44123 & house\_16H & $13,488$ & $16$ \\
      44125 & MagicTelescope & $13,376$ & $10$ \\
      \rowcolor{darkgray!15}44126 & bank-marketing & $10,578$ & $7$ \\
      45019 & Bioresponse & $3,434$ & $419$ \\
      \rowcolor{darkgray!15}44128 & MiniBooNE & $72,998$ & $50$ \\
      45020 & default-of-credit-card-clients & $13,272$ & $20$ \\
      \rowcolor{darkgray!15}44129 & Higgs  & $940,160$ & $24$ \\
      44130 & eye\_movements & $7,608$ & $20$ \\
      \rowcolor{darkgray!15}45022 & Diabetes130US & $71,090$ & $7$ \\
      45021 & jannis & $57,580$ & $54$ \\
      \rowcolor{darkgray!15}45089 & credit  & $16,714$ & $10$ \\
      45028 & california  & $20,634$ & $8$ \\
      \hline
      \end{tabular}
    \label{tab:num_class}
  \end{table*}

\begin{table*}[htbp]
  \centering
  \caption{Categorical regression}
    \small
    \begin{tabular}{||clrr||}
    \hline
    \rowcolor{darkgray!30}OpenML ID & Dataset Name & \#Samples & \#Features \\
    \hline
    \hline
    45041 & topo\_2\_1 & $8,885$ & $255$ \\
    \rowcolor{darkgray!15}44055 & analcatdata\_supreme & $4,052$ & $7$ \\
    44056 & visualizing\_soil & $8,641$ & $4$ \\
    \rowcolor{darkgray!15}45045 & delays\_zurich\_transport & $5,465,575$ & $12$ \\
    44059 & diamonds & $53,940$ & $9$ \\
    \rowcolor{darkgray!15}45046 & Allstate\_Claims\_Severity & $188,318$ & $124$ \\
    44061 & Mercedes\_Benz\_Greener\_Manufacturing & $4,209$ & $359$ \\
    \rowcolor{darkgray!15}44062 & Brazilian\_houses & $10,692$ & $11$ \\
    44063 & Bike\_Sharing\_Demand & $17,379$ & $11$ \\
    \rowcolor{darkgray!15}45047 & Airlines\_DepDelay\_1M & $1,000,000$ & $5$ \\
    44065 & nyc-taxi-green-dec-2016 & $581,835$ & $16$ \\
    \rowcolor{darkgray!15}45042 & abalone & $4,177$ & $8$ \\
    44066 & house\_sales & $21,613$ & $17$ \\
    \rowcolor{darkgray!15}45043 & seattlecrime6 & $52,031$ & $4$ \\
    45048 & medical\_charges & $163,065$ & $5$ \\
    \rowcolor{darkgray!15}44068 & particulate-matter-ukair-2017 & $394,299$ & $6$ \\
    44069 & SGEMM\_GPU\_kernel\_performance  & $241,600$ & $9$ \\
    \hline
    \end{tabular}
  \label{tab:cat_reg}
\end{table*}

\begin{table*}[htbp]
    \centering
    \caption{Categorical classification}
      \small
      \begin{tabular}{||clrr||}
      \hline
      \rowcolor{darkgray!30}OpenML ID & Dataset Name & \#Samples & \#Features \\
      \hline
      \hline
      44156 & electricity & $38,474$ & $8$ \\
      \rowcolor{darkgray!15}44157 & eye\_movements & $7,608$ & $23$ \\
      44159 & covertype & $423,680$ & $54$ \\
      \rowcolor{darkgray!15}45035 & albert & $58,252$ & $31$ \\
      45039 & compas-two-years & $4,966$ & $11$ \\
      \rowcolor{darkgray!15}45036 & default-of-credit-card-clients & $13,272$ & $21$ \\
      45038 & road-safety & $111,762$ & $32$ \\
      \hline
      \end{tabular}
    \label{tab:cat_class}
\end{table*}

\begin{table*}[htbp]
  \centering
  \caption{Image Classification}
    \small
    \begin{tabular}{||lcccc||}
    \hline
    \rowcolor{darkgray!30}Dataset & \#Train Samples & \# Test Samples & \#Categories & Input Size\\
    \hline
    \hline
    MNIST & $60,000$ & $10,000$ & $10$ & $28\times28$ \\
    \rowcolor{darkgray!15}Fashion-MNIST & $60,000$ & $10,000$ & $10$ & $28\times28$ \\
    K-MNIST & $60,000$ & $10,000$ & $10$ & $28\times28$ \\
    \rowcolor{darkgray!15}Q-MNIST & $60,000$ & $60,000$ & $10$ & $28\times28$ \\
    CIFAR-10 & $50,000$& 1$0,000$ & $10$ & $32\times32$ \\
    \rowcolor{darkgray!15}CIFAR-100 & $50,000$ & $10,000$ & $10$ & $32\times32$ \\
    \hline
    \end{tabular}
  \label{tab:img_class}
\end{table*}

\end{document}